\documentclass{article}

\usepackage{arxiv}
\usepackage{amsmath}
\usepackage[utf8]{inputenc} 
\usepackage[T1]{fontenc}    
\usepackage{hyperref}       
\usepackage{url}            
\usepackage{booktabs}       
\usepackage{amsfonts}       
\usepackage{nicefrac}       
\usepackage{microtype}      
\usepackage{lipsum}		
\usepackage{graphicx}
\usepackage{doi}
\usepackage{xcolor}
\usepackage{multirow}
\usepackage{subcaption}
\title{Towards In-Vehicle Multi-Task Facial Attribute Recognition: Investigating Synthetic Data and Vision Foundation Models}

\author{ \includegraphics[scale=0.06]{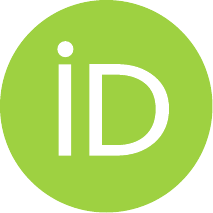}\hspace{1mm}Esmaeil Seraj\thanks{Corresponding author <email: esmaeil.seraj09@gmail.com>.} \\
	Research and Advanced Engineering\\
	2101 Village Rd, Room 3112\\
	Dearborn, MI 48124, USA \\
	\texttt{eseraj@ford.com} \\
	\And
	\includegraphics[scale=0.06]{orcid.pdf}\hspace{1mm}Walter Talamonti \\
	Research and Advanced Engineering\\
	2101 Village Rd, Room 3112\\
	Dearborn, MI 48124, USA \\
	\texttt{wtalamo1@ford.com} \\
}

\renewcommand{\shorttitle}{In-Vehicle Multi-Task Facial Attribute Recognition via Vision Foundation Models}

\hypersetup{
pdftitle={A template for the arxiv style},
pdfsubject={q-bio.NC, q-bio.QM},
pdfauthor={David S.~Hippocampus, Elias D.~Striatum},
pdfkeywords={First keyword, Second keyword, More},
}

\begin{document}
\maketitle

\begin{abstract}
	In the burgeoning field of intelligent transportation systems, enhancing vehicle-driver interaction through facial attribute recognition, such as facial expression, eye gaze, age, etc., is of paramount importance for safety, personalization, and overall user experience. However, the scarcity of comprehensive large-scale, real-world datasets poses a significant challenge for training robust multi-task models. Existing literature often overlooks the potential of synthetic datasets and the comparative efficacy of state-of-the-art vision foundation models in such constrained settings. This paper addresses these gaps by investigating the utility of synthetic datasets for training complex multi-task models that recognize facial attributes of passengers of a vehicle, such as gaze plane, age, and facial expression. Utilizing transfer learning techniques with both pre-trained Vision Transformer (ViT) and Residual Network (ResNet) models, we explore various training and adaptation methods to optimize performance, particularly when data availability is limited. We provide extensive post-evaluation analysis, investigating the effects of synthetic data distributions on model performance in in-distribution data and out-of-distribution inference. Our study unveils counter-intuitive findings, notably the superior performance of ResNet over ViTs in our specific multi-task context, which is attributed to the mismatch in model complexity relative to task complexity. Our results highlight the challenges and opportunities for enhancing the use of synthetic data and vision foundation models in practical applications.
\end{abstract}


\section{Introduction}
\label{sec:Introduction}
The advancement of intelligent transportation systems has opened new avenues for enhancing the interaction between vehicles and their drivers~\cite{fagnant2015preparing, tyagi2022autonomous, da2014artificial, li2022driver, natarajan2023human}. One such promising area is the use of in-cabin cameras to recognize facial attributes of the driver and passengers, thereby improving safety measures, personalizing user experiences, and facilitating more natural human-vehicle interactions~\cite{torstensson2019vehicle, kashevnik2019smartphone, sasidharan2015vehicle, kashevnik2020human, fernandez2016driver, seraj2019safe}.

However, the field faces several challenges that limit its potential. Existing literature primarily focuses on using real-world datasets for training models, which are often expensive and time-consuming to collect and can raise privacy concerns (i.e., collecting personal data for training models)~\cite{namazi2019using, chen2018real, ergenc2022review, torstensson2019vehicle, kashevnik2019smartphone}. Moreover, there is a lack of comprehensive studies that explore the efficacy of using synthetic datasets and pre-trained vision foundation models for in-vehicle facial recognition tasks. These gaps in the literature present a missed opportunity for leveraging alternative, possibly more efficient, methods for model training and evaluation such as the utility of synthetic data and pre-trained foundation models.

Existing machine learning methods for synthetic data generation span a variety of techniques, most notably Generative Adversarial Networks (GANs) and Variational Autoencoders (VAEs)~\cite{abufadda2021survey, lu2023machine, nikolenko2021synthetic}, as well as Diffisuion Models~\cite{gu2022vector, croitoru2023diffusion, ho2022cascaded}. These methods have been widely used to create high-fidelity, realistic synthetic datasets that can mimic the complexities of real-world data. The generated synthetic data serves as a valuable resource for training and validating machine learning models, especially in scenarios where collecting real-world data is challenging or costly, such as healthcare domains~\cite{chen2021synthetic, dahmen2019synsys} and autonomous driving~\cite{talwar2020evaluating, yuan2021comap}. Utilizing pre-trained models when working with limited synthetic data can offer significant advantages, such as reducing the training time and computational resources required~\cite{yosinski2014transferable}. Moreover, these models often come with learned features that can enhance the model's generalization capabilities, thereby improving performance on both in-distribution and out-of-distribution tasks~\cite{yosinski2014transferable}.
\begin{figure}[t!]
    \includegraphics[width=\linewidth]{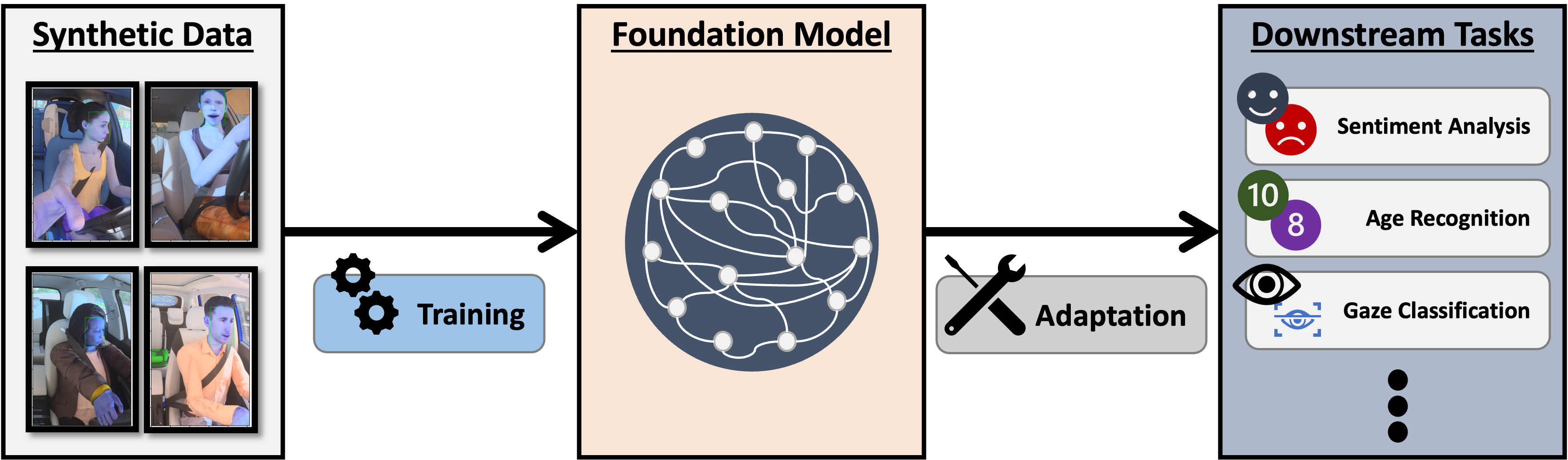}
    \caption{Example of a vision foundation model for in-vehicle perception and intelligence. Data is collected and used to train a deep neural network model. The trained model then generates a high-dimensional feature space representing the input data which can be used for any downstream task. To adapt the learned foundation model for downstream task, adaptation methods are needed.}
    \label{fig:vision_foundation_model}
    \vspace*{-0.25cm}
\end{figure}

The use of synthetic data and pre-trained models introduces its own set of challenges. Synthetic datasets, while easier to generate, may not perfectly replicate the complexities of real-world data~\cite{bach2015pixel, shorten2019survey, topol2019high}. Similarly, pre-trained models like Vision Transformers (ViTs)~\cite{dosovitskiy2021image, khan2022transformers, han2022survey} and Deep Residual Networks (ResNets)~\cite{he2016deep, Shafiq2022} come with their own limitations, such as robust adaptation, the risk of overfitting due to unbalanced model complexity with respect to task complexity, and the computational costs associated with their complexity~\cite{bommasani2021opportunities, mai2022towards, wang2023internimage}. These challenges necessitate a thorough investigation to ascertain the utility of such approaches in the context of multi-task facial attribute recognition in vehicles.

Motivated by these challenges and gaps, this paper aims to investigate the utility of synthetic datasets for training robust multi-task models capable of recognizing various facial attributes. We also explore the applicability of pre-trained vision foundation models (Figure~\ref{fig:vision_foundation_model}), specifically ViTs and ResNets, for this task. Additionally, we explore and test various adaptation methods such as linear probing, prefix tuning, and full fine tuning as well as different training techniques such as curriculum learning to improve model performance. We also perform several ablation studies and experiments on the effects of different preprocessing steps on model performances. Finally, we provide extensive post-evaluation analysis, investigating the effects of synthetic data distribution on model performance in in-distribution and out-of-distribution data. The goal is to provide a comprehensive analysis that not only evaluates the performance of these models but also delves into the nuances of training techniques and data quality.

\begin{figure}[t!]
    \centering
    \includegraphics[width=1\linewidth]{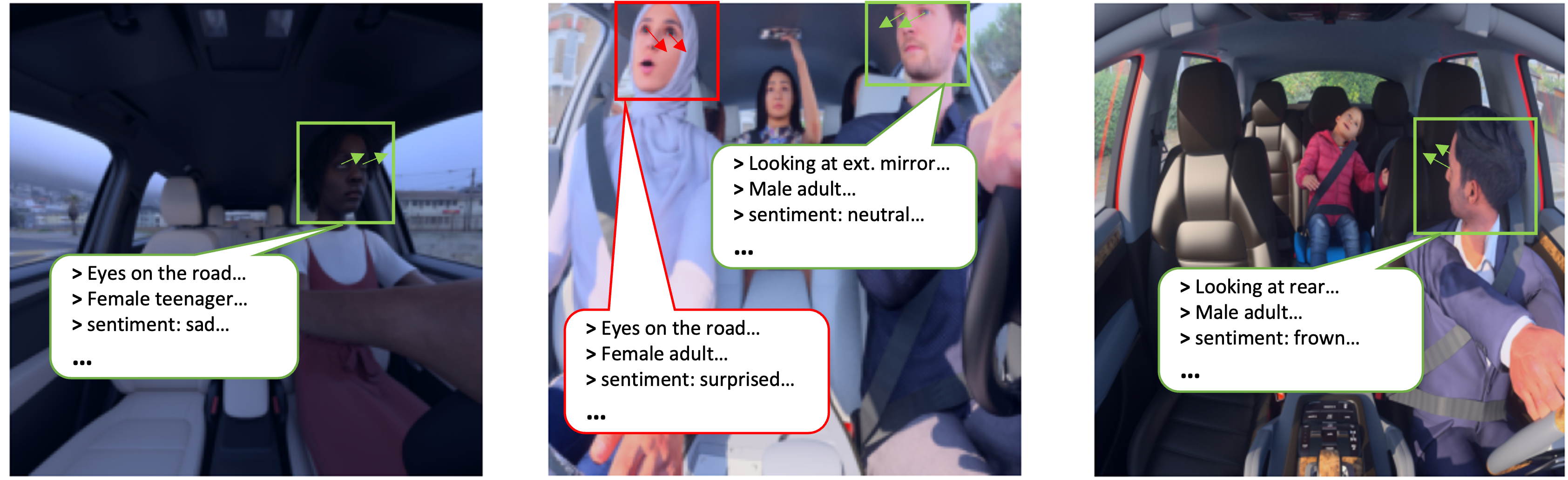}
    \caption{Example of in-vehicle perception and intelligence via multi-task facial attribute recognition on synthetically generated data. Figure demonstrates an in-cabin perception system capable of understanding several facial attributes, such as gaze, age, and facial expression of the driver and passengers. Such modular system which is built on readily generated synthetic data and existing vision foundation models can be adapted to any desired downstream task, enabling an enhanced vehicle-driver interaction and passenger experience.}
    \label{fig:example_incabin_ai}
    \vspace*{-0.25cm}
\end{figure}

In this work, we collect synthetic datasets from three major data generation companies. Herein, to protect proprietary and confidential information, we will refer to the three synthetic datasets as SynthA, SynthB, and SynthC datasets~\footnote{\textcolor{red}{Please note that, while the real dataset names are removed, to achieve reproducibility, we intend to release anonymized samples of each dataset in a public Github which will accompany the manuscript upon acceptance.}} (see Section~\ref{subsec:SyntheticDatasets} for more details about each dataset). We employ various training and adaptation techniques to train multi-task models empowered by existing vision foundation models. We particularly focus on building a perception model capable of understanding several facial attributes, such as human gaze, age, and facial expression (Figure~\ref{fig:example_incabin_ai}), of the vehicle driver or passengers. Such modular system which is built on synthetic data and existing vision foundation models can be adapted to any desired downstream task (i.e., tasks can readily be added or removed), enabling in-vehicle intelligence and enhanced vehicle-driver interaction. We delve into the intricacies of model architecture, layer configurations, and training regimes to provide a nuanced understanding of how these foundation models can be effectively adapted for automotive applications with available synthetic datasets. We evaluate the performance of these models using both in-distribution and out-of-distribution datasets. Our empirical evaluation focuses on a range of metrics, including but not limited to, data and label distributions, model performance, and data quality. Our key contributions and findings are as follows:
\begin{enumerate}
    \item A comprehensive study on the utility of synthetically generated datasets for training multi-task facial attribute recognition models and building perception models for in-vehicle intelligence.
    \item An in-depth analysis of the power, flexibility and performance of the pre-trained vision foundation models, revealing counter-intuitive findings such as the superior performance of ResNet over ViTs in our specific setting due to unbalanced model and task complexities.
    \item Exploring several training and adaptation techniques for vision foundation models in order to improve model performance, particularly in scenarios with limited data availability.
    \item Post-evaluation inference and data distribution analysis and providing critical insights into the limitations and potential improvements for using synthetic data in real-world applications.
\end{enumerate}

\section{Prior Work and Literature Review}
\label{sec:PriorWork}
The field of in-vehicle facial recognition has garnered significant attention in recent years, driven by the increasing integration of intelligent systems into automotive applications~\cite{murali2022intelligent, lu2019intelligent, archana2022face, fagnant2015preparing, tyagi2022autonomous, da2014artificial, torstensson2019vehicle, kashevnik2019smartphone, sasidharan2015vehicle}. This section provides an comprehensive overview of the existing literature, focusing on four main areas: intelligent and connected vehicles, driver-vehicle interactions, in-vehicle facial recognition technologies, and the use of vision foundation models in automotive settings. We also identify gaps in the current research landscape that our work aims to address.

\subsection{Intelligent and Connected Vehicles}
\label{subsec:IntelligentandConnectedVehicles}
In recent years, the field of intelligent and connected vehicles has garnered significant attention, serving as a cornerstone for the advancement of modern transportation systems~\cite{yang2018intelligent, mingyang2022survey, liu2018intelligent, seraj2023enhancing, chu2021cloud}. These vehicles are equipped with a myriad of sensors~\cite{xun2019experimental} and communication technologies~\cite{yang2018intelligent, mingyang2022survey, seraj2022embodied, seraj2022learning, liu2018intelligent, seraj2023embodied} that enable them to perceive their environment~\cite{du2022novel, du2020improved, du2019real, zhou2018velocity}, make decisions~\cite{okumura2016challenges, seraj2021heterogeneous, konan2023contrastive, seraj2023embodied, hu2021review, seraj2020firecommander}, control and maneuver~\cite{xu2021coordinated, seraj2020coordinated, xiaoping2020coordinated, seraj2021hierarchical, seraj2021adaptive}, and communicate with other vehicles and infrastructure~\cite{jia2016enhanced, seraj2022learning, dey2016vehicle, seraj2020firecommander, seraj2021heterogeneous, santa2008architecture, seraj2022embodied, konan2021iterated, seraj2023mixed, pimentel2022scaling}. The concept of Vehicle-to-Everything (V2X) communication has been extensively studied to facilitate real-time data exchange, enhancing road safety and traffic efficiency~\cite{jia2016enhanced, dey2016vehicle, santa2008architecture}. Machine Learning (ML) and Computer Vision (CV) techniques have been employed for various in-vehicle perception tasks, such as driver monitoring~\cite{xun2019experimental, tan2021human, murali2022intelligent}, object detection~\cite{cafiso2017vehicle, dinakaran2020vehicle, huang1998object, qi2019convolutional, murali2022intelligent}, and path planning~\cite{aradi2020survey, yudin2019object, xie2021unmanned}.

\subsection{Driver-Vehicle Interactions}
\label{subsec:Driver-VehicleInteractions}
In the realm of driver-vehicle interactions~\cite{tan2021human, murali2022intelligent}, a plethora of research has been conducted to explore various facets of this complex relationship, with the overarching goal of enhancing safety~\cite{hu2021review}, efficiency~\cite{tan2021human, yang2018intelligent}, and user experience~\cite{normark2021personalizable, du2023chat, watsonambient, uvarov2022maintaining, hasenjager2017personalization, xu2021research}. Studies have delved into the use of advanced Human-Machine Interfaces (HMIs), such as touchscreens, voice commands~\cite{du2023chat}, and even gesture recognition~\cite{jacob2015hand, pickering2005gesture, agrawal2013emotion}, to facilitate seamless communication between the driver and the vehicle. Additionally, there has been a growing interest in leveraging machine learning algorithms to understand and predict driver behavior~\cite{li2022driver, doshi2011tactical, kang2013various, du2022novel}, including attention level~\cite{yang2020all, hu2022novel, rong2022and}, emotional state~\cite{agrawal2013emotion}, and intent\cite{zyner2017long, morris2011lane, phillips2017generalizable}. These predictive models often utilize data from in-car sensors, cameras, and other IoT devices to provide real-time feedback or even automated interventions, such as adaptive cruise control adjustments or lane-keeping assistance~\cite{morris2011lane, du2022novel}. However, much of the existing work has relied on traditional data collection methods and single-task learning models, overlooking the potential of multi-task learning frameworks and synthetic datasets, which are the focus of our paper. Our research aims to extend the current understanding by exploring the efficacy of multi-task facial attribute recognition models in enhancing driver-vehicle interactions, particularly when trained on synthetic data.

\subsection{In-Vehicle Facial Recognition}
\label{subsec:InVehicleFacialRecognition}
In-vehicle facial recognition technologies have primarily been developed to enhance safety and user experience~\cite{hu2021review, jia2016enhanced, dey2016vehicle, santa2008architecture, torstensson2019vehicle, kashevnik2019smartphone, sasidharan2015vehicle}. These systems often employ a variety of machine learning algorithms, including Convolutional Neural Networks (CNNs)~\cite{li2021survey, o2015introduction}, to detect and analyze facial attributes such as gaze direction~\cite{vicente2015driver, yoon2019driver, naqvi2018deep, fletcher2009driver, kang2013various, shah2022driver, fridman2016driver, dari2020neural}, age~\cite{wang2015deeply, ashiqur2021deep, lee2018applying}, and emotional state~\cite{du2020convolution, zepf2020driver, li2021cogemonet, xiao2022road, agrawal2013emotion}. The majority of these studies rely on real-world datasets, which are labor-intensive and costly to collect. Moreover, these datasets often suffer from issues of imbalance and lack of diversity, which can lead to biased or less robust models.

The existing literature largely overlooks the potential of synthetic datasets for training ML and CV models~\cite{abufadda2021survey, lu2023machine, nikolenko2021synthetic}. The focus has been on traditional data collection methods, such as in-car cameras and mobile devices, which may not be scalable or cost-effective for broader applications and could raise privacy concerns (i.e., collecting personal data for training models)~\cite{talwar2020evaluating, yuan2021comap}. Our work aims to fill this gap by investigating the utility of synthetic datasets for training multi-task facial recognition models in vehicles. We also explore the use of advanced transfer learning and adaptation techniques to improve the generalizability and robustness of these models, empowered by existing pre-trained foundation models. The incorporation of synthetic data not only alleviates the need for extensive real-world data collection but also provides a controlled environment to simulate various lighting conditions, angles, and occlusions that are critical for in-vehicle applications~\cite{abufadda2021survey, lu2023machine, nikolenko2021synthetic}. Overall, our work contributes to the growing body of research that advocates for a more flexible and scalable approach to data collection and model training in the realm of in-vehicle facial recognition. 

\subsection{Vision Foundation Models for Automotive Applications}
\label{subsec:VisionFoundationModels}
Vision foundation models like Convolutional Neural Networks (CNNs)~\cite{li2021survey, o2015introduction}, ResNets~\cite{he2016deep, Shafiq2022}, and more recently, Vision Transformers (ViTs)~\cite{dosovitskiy2021image, khan2022transformers, han2022survey}, have been employed in various automotive applications, ranging from object detection~\cite{cafiso2017vehicle, dinakaran2020vehicle, huang1998object, qi2019convolutional, murali2022intelligent} to lane tracking~\cite{morris2011lane, du2022novel}, and even driver monitoring systems~\cite{li2022driver, doshi2011tactical, kang2013various, du2022novel}. These models offer robust performance but often require substantial computational resources~\cite{park2022fast, liu2022video, liu2021swin}, making them less ideal for real-time, in-vehicle applications. The computational burden is further exacerbated when these models are deployed in embedded systems with limited processing capabilities, which are common in automotive settings.

A limitation in the current literature is the lack of comprehensive studies that explore the trade-offs between different types of vision foundation models, especially in the context of multi-task learning for in-vehicle applications. This gap is significant because each type of foundation model has its own set of advantages and limitations that could be more or less suitable for specific tasks. For instance, CNNs are generally good for spatial hierarchies but may struggle with long-range dependencies~\cite{donahue2015long, vaswani2017attention, gehring2017convolutional, hosseini2017limitation, yamashita2018convolutional, nebauer1998evaluation, li2018survey, wang2020recurrent}, whereas ViTs excel in capturing such dependencies but might be computationally more intensive~\cite{vaswani2017attention, khan2022transformers, islam2022recent, yuan2021tokens, thisanke2023semantic, long2023lacvit, li2021localvit, xu2023vision}. Furthermore, there is minimal exploration of training techniques that could optimize these models for better performance and efficiency, such as fine-tuning, transfer learning, and data augmentation strategies tailored for automotive scenarios.

Our work addresses these limitations by evaluating the performance of pre-trained ResNets and ViTs in a multi-task learning framework, while also exploring various training and adaptation techniques. We delve into the intricacies of model architecture, layer configurations, and training regimes to provide a nuanced understanding of how these foundation models can be effectively adapted for automotive applications. Additionally, we investigate the impact of using synthetic data to train these models, offering insights into how the quality and distribution of training data can affect both in-distribution and out-of-distribution performance. Overall, our research contributes to a more holistic understanding of the role and optimization of vision foundation models in the rapidly evolving field of automotive intelligence, particularly when paired with limited amount, synthetically generated datasets.

\section{Background}
\label{sec:Background}
This section provides a comprehensive background on the key concepts and methodologies that underpin our research. We delve into multi-task learning~\cite{zhang2018overview}, transfer learning~\cite{weiss2016survey, torrey2010transfer}, and vision foundation models~\cite{bommasani2021opportunities}, elucidating the mathematical foundations and practical implications of each.

\subsection{Multi-Task Learning}
\label{subsec:MultiTaskLearning}
Multi-Task Learning (MTL) is a learning paradigm where a single model is trained to perform multiple tasks simultaneously~\cite{zhang2018overview, ruder2017overview, zhang2021survey, thung2018brief}. The primary advantage of MTL is that it allows the model to leverage shared representations across tasks, often leading to improved generalization~\cite{evgeniou2004regularized, standley2020tasks, nguyen2019multi}. The general loss function for an MTL model can be formulated as in Equation~\ref{eq:MTL_loss}, where $\mathcal{L}_i(\theta)$ is the loss for task $i$, $w_i$ is the weight for task $i$, $\theta$ are the model parameters, and $N$ is the number of tasks~\cite{zhang2018overview}.
\begin{equation}
    \label{eq:MTL_loss}
    \mathcal{L}(\theta) = \sum_{i=1}^{N} w_i \mathcal{L}_i(\theta)
\end{equation}

In the context of our research, MTL is particularly relevant for training a single model on top of existing powerful vision foundation models to recognize multiple downstream tasks, i.e., facial attributes, thereby reducing the computational load and increasing the efficiency of in-vehicle systems.

\subsection{Transfer Learning}
\label{subsec:TransferLearning}
Transfer Learning (TL) is a machine learning technique where a model developed for a particular task is adapted for a second related task~\cite{weiss2016survey, torrey2010transfer, zhuang2020comprehensive, niu2020decade, pan2009survey}. In the context of neural networks, this often involves fine-tuning a pre-trained model on a new dataset~\cite{yang2020transfer, li2020transfer, wu2021research, gopalakrishnan2017deep}. The loss function for transfer learning can be expressed as in Equation~\ref{eq:TL_loss}, where $\mathcal{L}_{\text{source}}(\theta)$ is the loss on the source task, $\mathcal{L}_{\text{target}}(\theta')$ is the loss on the target task, $\theta$ and $\theta'$ are the model parameters for the source and target tasks respectively, and $\lambda$ is a regularization term~\cite{weiss2016survey, torrey2010transfer}.
\begin{equation}
    \label{eq:TL_loss}
    \mathcal{L}(\theta, \theta') = \mathcal{L}_{\text{source}}(\theta) + \lambda \mathcal{L}_{\text{target}}(\theta')
\end{equation}

Transfer learning is pivotal in our work for adapting existing pre-trained vision foundation models, such as ViTs and ResNets, in order to exploit their power and flexibility for our specific multi-task learning problem, thereby saving time and computational resources.

\subsection{Vision Foundation Models}
\label{subsec:VisionFoundationModels}
Foundation models serve as a cornerstone in the machine learning landscape, offering pre-trained architectures that can be fine-tuned for a wide array of specific tasks~\cite{bommasani2021opportunities}. These models are trained on large, diverse datasets, enabling them to capture intricate patterns and relationships in the data~\cite{awais2023foundational, bommasani2021opportunities}. When specialized for computer vision tasks, these foundation models are referred to as vision foundation models~\cite{awais2023foundational, zhang2023survey, zhang2023vision}. They have been instrumental in advancing the field of computer vision, offering robust performance across a range of tasks from object detection to semantic segmentation~\cite{awais2023foundational, zhang2023survey}. The adaptability and generalizability of these models make them a valuable asset in the development of specialized computer vision applications.

In the context of our research, vision foundation models play a pivotal role in achieving high-performance multi-task learning for in-vehicle facial attribute recognition. Leveraging pre-trained vision foundation models such as ResNet and ViTs allows us to bypass the need for extensive data collection and training from scratch. This is particularly beneficial given the resource constraints and the need for real-time processing in automotive applications. By fine-tuning these models on our specific tasks via advanced adaptation techniques (Figure~\ref{fig:vision_foundation_model}), we can achieve enhanced performance while significantly reducing the time and computational resources required for model development. In the subsequent sections, we will provide a succinct overview of two foundational models integral to our research: Residual Networks (ResNet) and Vision Transformers (ViT). ResNet models are the most prevalent foundation models and have established themselves as a cornerstone in the realm of computer vision tasks, while ViTs represent the latest advancements in architectural design, garnering significant attention for their exceptional computational efficacy and performance metrics.

\subsubsection{Residual Network (ResNet) Model}
\label{subsubsec:ResNetModel}
Residual Networks (ResNets)~\cite{he2016deep, Shafiq2022} are a type of Convolutional Neural Network (CNN)~\cite{li2021survey, o2015introduction} that include skip connections to allow gradients to flow through the network more easily. This architecture mitigates the vanishing gradient problem, making it possible to train very deep networks effectively.

The defining feature of ResNets is the incorporation of residual connections, also known as skip connections, that bypass one or more layers~\cite{he2016deep, Shafiq2022}. These connections are added back to the output of the stacked layers, forming the final output of the residual block. Mathematically, this can be represented as $F(x) = H(x) - x$, where $F(x)$ is the residual mapping to be learned and $H(x)$ is the original mapping~\cite{he2016deep, Shafiq2022}. The residual connections enable the backpropagation of gradients all the way through the network, mitigating the vanishing gradient problem commonly encountered in deep networks. In terms of spatial feature learning, the residual layers allow ResNets to learn more complex representations by combining both local and global contextual information from different layers, thereby enhancing the network's ability to recognize intricate patterns in images~\cite{he2016deep, Shafiq2022}.

\subsubsection{Vision Transformer (ViT)}
\label{subsubsec:VisionTransformers}
Vision Transformers (ViTs) leverage the Transformer architecture, originally designed for natural language processing tasks, for computer vision applications~\cite{dosovitskiy2021image, khan2022transformers, han2022survey}. Unlike CNNs, ViTs divide an image into patches and process them in parallel through the Transformer layers~\cite{vaswani2017attention, khan2022transformers, islam2022recent}. This approach allows for more global feature extraction, making ViTs particularly effective for complex vision tasks.

ViTs are a novel class of models that adapt the Transformer architecture~\cite{vaswani2017attention}, originally designed for natural language processing, to the realm of computer vision. The architecture consists of multiple transformer layers, each comprising multi-head self-attention mechanisms and feed-forward neural networks~\cite{dosovitskiy2021image, khan2022transformers, han2022survey}. The attention mechanism is particularly noteworthy for its ability to capture long-range dependencies in the data~\cite{vaswani2017attention}. In the context of images, this means that ViTs can learn spatial hierarchies and relationships between distant pixels, thereby capturing more global features. The attention mechanism works by computing a weighted sum of all pixels, where the weights are determined by the similarity between the query, key, and value representations of each pixel. To perform these processes, however, an image must be \textit{patchified} (i.e., cropped and converted into a series of sequential patches). This is similar to turning an image into several word-embeddings in a sentence. This enables the model to focus on different parts of the image adaptively, providing a more nuanced understanding of the spatial features present. The aforementioned process in ViTs are mathematically described in Equations~\ref{eq:patchification}-\ref{eq:transformer}.
\begin{equation}
    \label{eq:patchification}
    \text{Patch Embedding: } p_i = \text{Conv2D}\left(\text{Linear}(x_i)\right)
\end{equation}
\begin{equation}
    \label{eq:transformer}
    \text{Global Feature: } Z = \text{Transformer}(p_1, p_2, \ldots, p_n)
\end{equation}

\section{Methodology}
\label{sec:Methodology}
This section delineates the comprehensive methodology employed in our research to investigate the utility of synthetic data and pre-trained vision foundation models for multi-task learning in in-vehicle facial attribute recognition. The section is organized into several subsections, each focusing on a critical aspect of our research pipeline. We begin with the preprocessing steps (Section~\ref{subsec:Preprocessing}), detailing the procedures for data cleaning, normalization, and preparation. Subsequent subsections delve into the architecture of the multi-task models (Section~\ref{subsec:MultiTaskModelArchitecture}), the adaptation techniques used for transfer learning (Section~\ref{subsec:Adaptation}), and the specific training and evaluation strategies (Section~\ref{subsec:TrainingAndEvaluation}). The section aims to provide a thorough understanding of the experimental setup, thereby enabling replicability and further exploration.

\subsection{Preprocessing}
\label{subsec:Preprocessing}
In our research, we utilized two separate preprocessing pipelines, where the efficacy and performance of each will be rigorously evaluated in the results section. The first pipeline utilizes the original full images captured by in-cabin cameras. The second pipeline, on the other hand, extracts the face of the driver/passengers and uses face-only images for training the models. To extract the human faces in the second pipeline, we tested two prevalent face-extraction pre-trained models: (1) OpenCV's Internal Haar Cascade pre-trained model~\cite{cuimei2017human} and (2) LightFace's RetinaFace pre-trained model~\cite{serengil2020lightface, serengil2021lightface}. Our experiments indicate that the RetinaFace model significantly outperforms the Haar Cascade model across various lighting conditions, head positions, head sizes, and face-background color contrasts. As such, our presented results for face-only data in Section~\ref{sec:EmpiricalEvaluation} utilize the RetinaFace model. Both preprocessing pipelines (i.e., full-image data and face-only data) share the following steps: 
\begin{enumerate}
    \item \textbf{Normalization}: Normalization is used to adjust the data distribution to match that of the pre-trained models. Both ResNet and ViT foundation models employed in our work are trained on the ImageNet dataset~\cite{deng2009imagenet}. Therefore, the distribution of our data is normalized to fit that of the ImageNet dataset.
    \item \textbf{Downsampling and Resizing}: Downsampling and Resizing of images are used to fit the input shape of the pre-trained models. For instance, the pre-trained ViT b\_16 model works with input images of size $224\times224$.
    \item \textbf{Scaling}: Data scaling to transform the RGB range from $0-255$ to $0-1$. This is a common preprocessing step in computer vision models to help achieve numerical stability, faster Convergence, and improved generalization.
    \item \textbf{Shuffling}: Shuffling of the dataset to introduce randomness and improve generalization.
    \item \textbf{Splitting}: Train-test splitting, allocating $70$\% of the data for training and $30$\% for testing.
\end{enumerate}

We did not employ any data augmentation techniques. Instead, we conducted experiments on each synthetic dataset separately and also on a combined dataset comprising all three synthetic datasets, i.e., SynthA, SynthB, and SynthC (see Section~\ref{subsec:SyntheticDatasets}). Additionally, specific preprocessing steps were required for each synthetic dataset, as described below. Further details regarding each dataset, provided labels, and categories are provided in Section~\ref{subsec:SyntheticDatasets}:
\begin{itemize}
    \item In the \textit{SynthA} dataset, the age labels are provided as a numerical value, which is ambiguous and unhelpful for age recognition task. As such, the numerical age labels were transformed into age ranges as follows: \texttt{\{`0-3', `4-12', `13-18', `19-30', `31-50', `50+'\}}.
    
    \item In the \textit{SynthB} dataset some images, some labels were missing, despite a human driver was present in the respective image (and vice versa), and therefore, an additional cleaning step was necessary to remove data with problematic labels or images.
    
    \item For the \textit{SynthC} dataset, we re-annotated the gaze labels, reducing the number of originally provided gaze planes from 17 to seven based on our specific application-dependent interests.
\end{itemize}

\subsection{Multi-Task Model Architecture}
\label{subsec:MultiTaskModelArchitecture}
\begin{figure}[h!]
    \centering
    \includegraphics[width=1\linewidth]{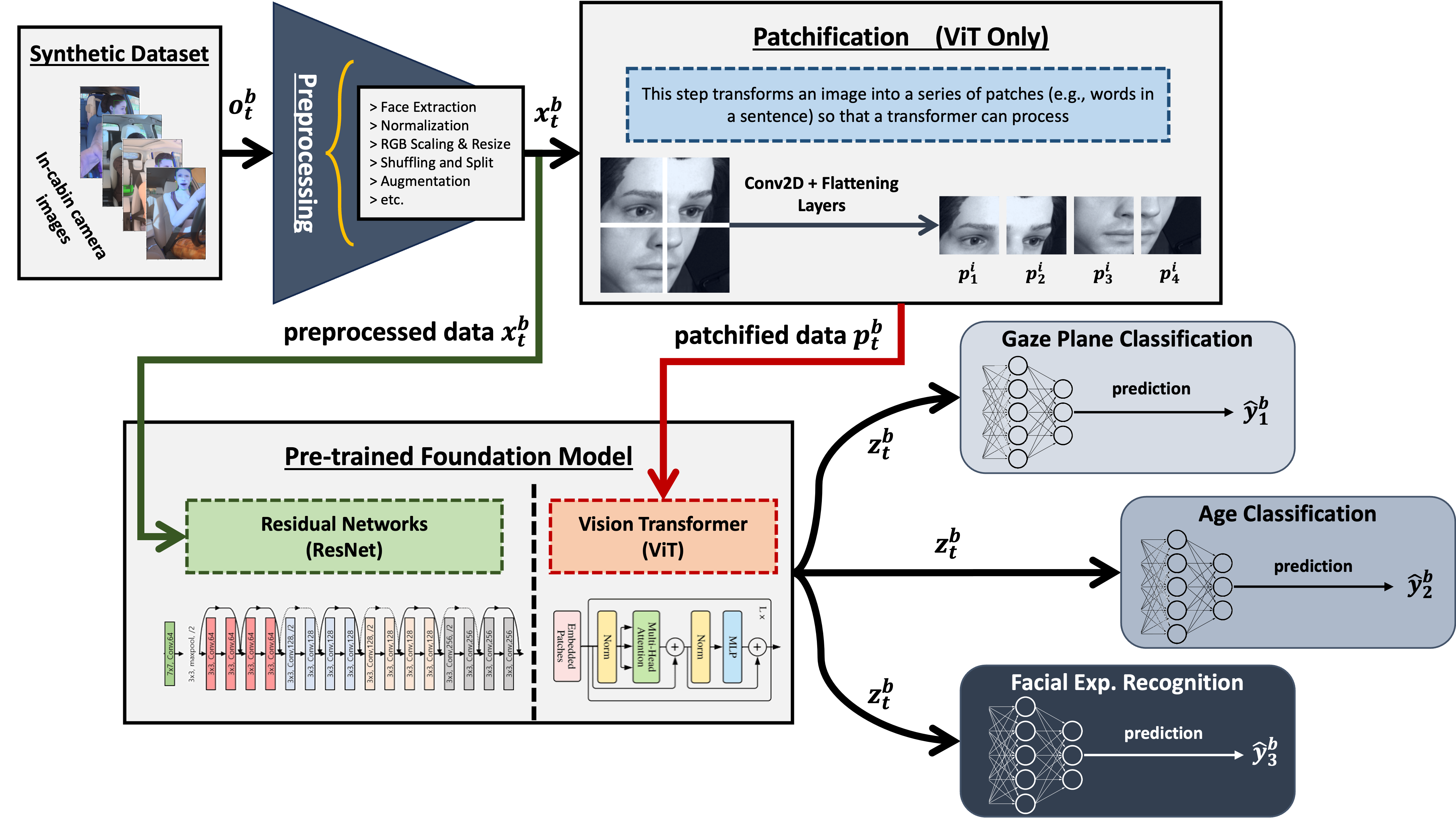}
    \caption{Our employed multi-task facial attribute recognition architecture via transfer learning from pre-trained vision foundation models. Our research employs two distinct vision foundation models as the backbone for our multi-task learning architecture: the Vision Transformer (ViT) and the Residual Network (ResNet). Both architectures share a common preprocessing block. The patchification block is only applied for the ViT model and is crucial for adapting image data into a format that the transformer architecture can effectively process. The output from these pre-trained models, which represents a high-dimensional feature space, is then directed into three separate task heads. The final layer in each separate task head is a linear layer with number of neurons equal to the number of classes for the respective task, outputting a probability vector for each prediction.}
    \label{fig:architecture}
    \vspace*{-0.5cm}
\end{figure}

Our multi-task model architecture for facial attribute recognition is shown in Figure~\ref{fig:architecture}. We employ two distinct vision foundation models as the backbone for our multi-task learning architecture: the Vision Transformer (ViT) and the Residual Network (ResNet). Both architectures share a common preprocessing block that performs the operations outlined in the previous subsection~\ref{subsec:Preprocessing}. However, the ViT model incorporates an additional \textit{Patchification} block (i.e., cropping and converting an image into a series of sequential patches). As shown in Equations~\ref{eq:patchification}-\ref{eq:transformer}, the patchification block employs a Conv2D layer to transform a $224\times224$ image into a series of $16\times16$ patches. These patches are subsequently flattened and stacked, serving as the input to the transformer layers in the ViT model. This patchification step is crucial for adapting image data into a format that the transformer architecture can effectively process.

For the ResNet model, the patchification step is omitted, and the output from the preprocessing block directly feeds into the pre-trained model (Figure~\ref{fig:architecture}). We utilized the ViT b\_16 and ResNet 18 variants for our experiments. The ViT b\_16 model consists of 12 transformer layers, each with multi-head self-attention mechanisms and feed-forward neural networks, as detailed in Subsection \ref{subsubsec:VisionTransformers}. On the other hand, the ResNet 18 model is composed of 18 layers, including multiple residual blocks that leverage skip connections to facilitate the training of deeper networks, as described in Subsection \ref{subsubsec:ResNetModel}. Each model offers unique advantages: the ViT is renowned for its superior performance in complex vision tasks, while the ResNet provides a more lightweight architecture that is well-suited for edge deployments.

As shown in Figure~\ref{fig:architecture}, the output from these pre-trained models, which represents a high-dimensional feature space, is then directed into three separate task heads. Each task head is responsible for one of the facial attributes we aim to predict: (1) gaze plane, (2) age, and (3) facial expression. These task heads are structurally identical, consisting of multiple fully-connected layers, dropout layers for regularization, and ReLU activation functions. The final layer in each task head is a linear layer with number of neurons equal to the number of classes for the respective task, outputting a probability vector for each prediction.

The architecture is designed to be highly modular, particularly at the task heads. This modularity allows for easy adaptability and experimentation with different tasks without requiring changes to the underlying foundation model. It also facilitates the incorporation of additional tasks in the future, thereby enhancing the model's utility across a broader range of applications. Full schematic representation of the architecture is provided in Figure~\ref{fig:architecture} for further clarity.

\subsection{Adaptation}
\label{subsec:Adaptation}
In order to adapt the pre-trained vision foundation models to our specific multi-task learning problem, we employed three distinct adaptation methods: linear probing, prefix tuning, and full fine-tuning~\cite{tan2018survey, weiss2016survey, pan2009survey}. Each of these methods offers a unique approach to leveraging the learned feature spaces of the pre-trained models for our specific tasks, and they come with their own sets of advantages and disadvantages.

\textbf{Linear Probing (LP):} In this method, the pre-trained vision foundation models are kept frozen during the training process, serving solely as feature extractors. The high-dimensional feature space generated by these models is then fed into the task heads, which are the only components updated via gradient descent. This approach is computationally efficient and straightforward, but its performance is highly dependent on the quality of the pre-trained model. It assumes that the feature space learned during pre-training is sufficiently expressive for the downstream tasks, which may not always be the case~\cite{tan2018survey, weiss2016survey, pan2009survey}.

\textbf{Prefix Tuning (PT):} This method unfreezes a small portion of the pre-trained vision foundation model, allowing it to be updated alongside the task heads during training. For the ViT model, we unfroze the last encoder layer, while for the ResNet model, the last residual block was made trainable. This selective fine-tuning enables the feature space to adapt more closely to the specific tasks at hand. Prefix tuning strikes a balance between computational efficiency and performance, often outperforming linear probing, especially when the dataset size is moderate~\cite{li2021prefix, tan2018survey, weiss2016survey, pan2009survey}.

\textbf{Full Fine-Tuning (FFT):} In this approach, the entire architecture, including the pre-trained vision foundation model, is updated via gradient descent. The pre-trained models essentially serve as an initialization point or a "warm-start" which facilitates faster convergence. This method is the most computationally intensive but also the most effective when abundant data and computational resources are available. It allows the model to fully adapt to the specific tasks, often resulting in the highest performance metrics among the three methods~\cite{tan2018survey, weiss2016survey, pan2009survey}.

Each of these adaptation methods has its own trade-offs. LP is computationally less demanding but may suffer from performance limitations. PT offers a middle ground, being more adaptive than linear probing while being less resource-intensive than full fine-tuning~\cite{li2021prefix}. FFT, although computationally expensive, often yields the best performance, making it the method of choice when resources are not a constraint. A comparative illustration of these adaptation methods is provided in Figure X for further clarity~\cite{li2021prefix, tan2018survey, weiss2016survey, pan2009survey}.

The efficacy of these adaptation methods, i.e., LP, PT, and FFT, is not solely determined by computational resources or the volume of available data; it is also intricately linked to the data distribution and the complexity of the downstream tasks. For instance, FFT may not always be the optimal choice for out-of-distribution (OOD) data, especially when the pre-trained features are of high quality~\cite{kumar2022fine, li2021prefix}. In such cases, fine-tuning the entire model can lead to a degradation in performance due to distributional shifts~\cite{kumar2022fine}. On the other hand, the complexity of the downstream tasks also plays a crucial role in the selection of the adaptation method. For relatively simpler tasks, extensive fine-tuning, as in FFT, can result in overfitting, thereby negatively impacting the model's generalization capabilities. Therefore, the choice of adaptation strategy should be made judiciously, taking into account not just the available resources and data, but also the specific characteristics of the tasks and the data distribution~\cite{kumar2022fine, li2021prefix, bommasani2021opportunities}.

\subsection{Training and Evaluation}
\label{subsec:TrainingAndEvaluation}
The training and evaluation of our multi-task model were conducted in a unified framework, leveraging a composite loss function and multiple evaluation metrics. The optimization process was designed to be both robust and efficient, incorporating various techniques to ensure the model's performance across different tasks.

\paragraph{Loss Function}
The loss function is a critical component in training neural networks, and in our case, it is a composite function that combines the Cross Entropy Loss (CEL) for each of the three task heads, i.e., gaze, age, and facial expression. Mathematically, the loss function $\mathcal{L}$ can be represented as in Equation~\ref{eq:our_loss}, where $y_{i}$ and $\hat{y}_{i}$ are the true and predicted labels for each task, and $\lambda_{i}$ is the weight decay for L2 regularization specific to each task head.:
\begin{equation}
    \label{eq:our_loss}
    \mathcal{L} = \sum_{i=1}^{3} \left( \text{CrossEntropy}(y_{i}, \hat{y}_{i}) + \lambda_{i} \cdot \text{L2 regularization term} \right)
\end{equation}

\paragraph{Optimization}
For the optimization process, we employed the Adam optimizer~\cite{kingma2014adam}, a widely-used optimization algorithm that combines the advantages of both AdaGrad~\cite{duchi2011adaptive} and RMSProp~\cite{ruder2016overview}. Adam is particularly effective in handling non-convex optimization landscapes, a common challenge in multi-task learning. Its adaptive learning rate capabilities make it a robust choice for our complex model architecture~\cite{kingma2014adam, ruder2016overview}.

\paragraph{Evaluation Metrics}
Both loss and accuracy metrics were used for evaluation. Specifically, we monitored the total loss and accuracy, as well as task-specific losses and accuracies. The model was trained and evaluated in the same loop over a batch of input data, iterating through multiple epochs. The model with the lowest evaluation loss was saved for inference, obviating the need for early stopping.

\paragraph{Training Techniques}
Several specific training techniques were employed to enhance the model's performance. First, we used learning rate schedules, starting with an initial learning rate of $lr = 1 \times 10^{-3}$ and decaying it by a factor of 0.5, with a minimum allowed learning rate of $1 \times 10^{-6}$. Second, we used empirically chosen loss weights for each task to balance their contributions to the overall loss. Lastly, we experimented with curriculum learning, where task losses were added incrementally based on their evaluation loss. Although this technique did not make it into all our experiments, its impact will be discussed as part of an ablation study in the results Section~\ref{sec:EmpiricalEvaluation}.

\section{Empirical Evaluation}
\label{sec:EmpiricalEvaluation}
In this comprehensive section, we present an empirical evaluation of our multi-task learning model, focusing on its performance across various synthetic datasets and under different experimental conditions. We initiate the discussion with an in-depth examination of the synthetic datasets used for training and validation (Section~\ref{subsec:SyntheticDatasets}). Following this, we delve into the model's performance on these individual datasets as well as a combined dataset, providing learning curves and task-specific metrics for both ViT and ResNet architectures (Section~\ref{subsec:ModelPerformances}). Ablation studies are then presented to assess the impact of different adaptation methods and curriculum learning strategies (Section~\ref{subsec:Ablation_Studies}). We also include a section dedicated to the model's inference performance on out-of-distribution (OOD) data (Section~\ref{subsec:OOD_Results}). Finally, we offer a post-evaluation study, analyzing the data distributions alongside achieved model performances, and provide insights into the observed results (Section~\ref{subsec:Discussion}). This section aims to provide a thorough understanding of the model's capabilities and limitations when paired with synthetic data.
\begin{figure}[t!]
    \centering
    \includegraphics[width=1\linewidth]{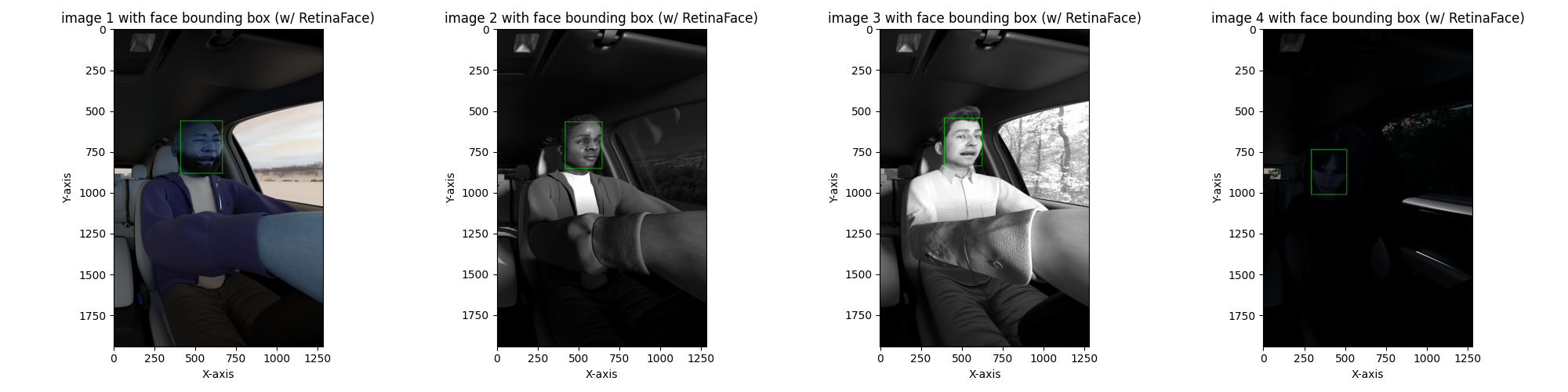}
    \caption{Image samples from the SynthA dataset. The RetinaFace pre-trained model~\cite{serengil2020lightface, serengil2021lightface} has been applied to extract the face bounding box (i.e., green box) of the driver.}
    \label{fig:SynthesisAI_sample_faces_retina}
\end{figure}

\subsection{Synthetic Datasets}
\label{subsec:SyntheticDatasets}
In this subsection, we provide an in-depth analysis of the synthetic datasets employed in our study. These datasets were generated by different sources and are designed to simulate various real-world scenarios involving in-cabin cameras capturing drivers and passengers. We focus on three primary datasets: SynthA, SynthB, and SynthC~\footnote{\textcolor{red}{Please note that er replaced the real names of the synthetic data generation companies due to proprietary an confidential information. However, to achieve reproducibility, we intend to release anonymized samples of each dataset in a public Github which will accompany the manuscript upon acceptance.}}. Each dataset has its unique characteristics, which we outline below.

\paragraph{SynthA:}
The SynthA dataset consists of 10,000 images captured by an in-cabin camera focusing on a driver in various driving scenarios, such as different lighting conditions and driver positions. The images in this dataset are frames extracted from short videos, which results in relatively low diversity compared to the other datasets. Samples of the SynthA dataset are demonstrated in Figure~\ref{fig:SynthesisAI_sample_faces_retina}.

\begin{figure}[t!]
    \centering
    \includegraphics[width=1\linewidth]{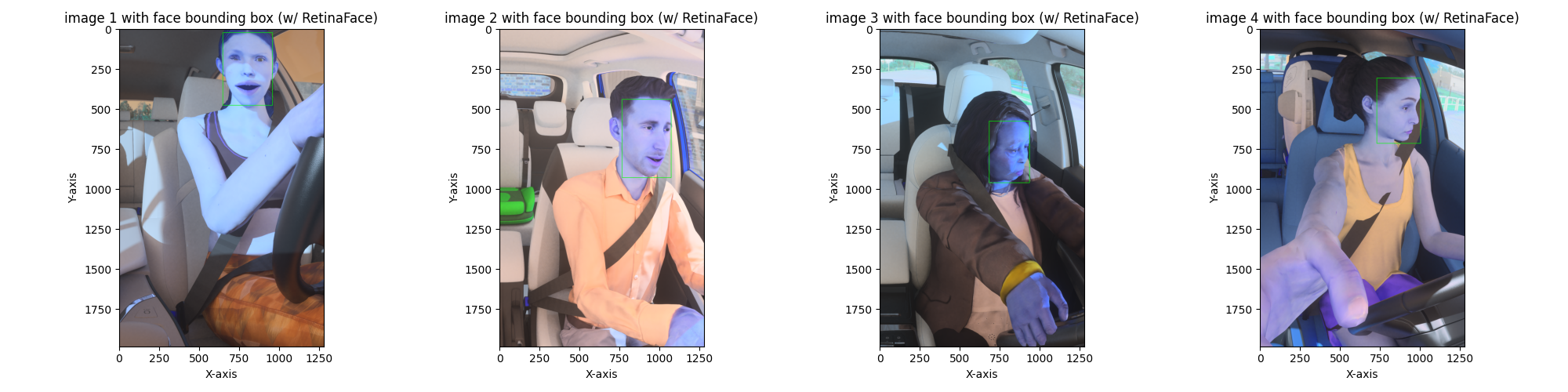}
    \caption{Image samples from the SynthB dataset. The RetinaFace pre-trained model~\cite{serengil2020lightface, serengil2021lightface} has been applied to extract the face bounding box (i.e., green box) of the driver.}
    \label{fig:Anyverse_sample_faces_retina}
    \vspace*{-0.5cm}
\end{figure}

\paragraph{SynthB:}
The SynthB dataset is more extensive, containing 2,999 images (post-cleaning). These images capture multiple passengers and a driver in a wide array of scenarios, including different lighting conditions and positions. The dataset is highly diverse, featuring drivers and passengers of various races and ages with different head- or face-wears. Samples of the SynthB dataset are demonstrated in Figure~\ref{fig:Anyverse_sample_faces_retina}.

\begin{figure}[h!]
    \centering
    \includegraphics[width=1\linewidth]{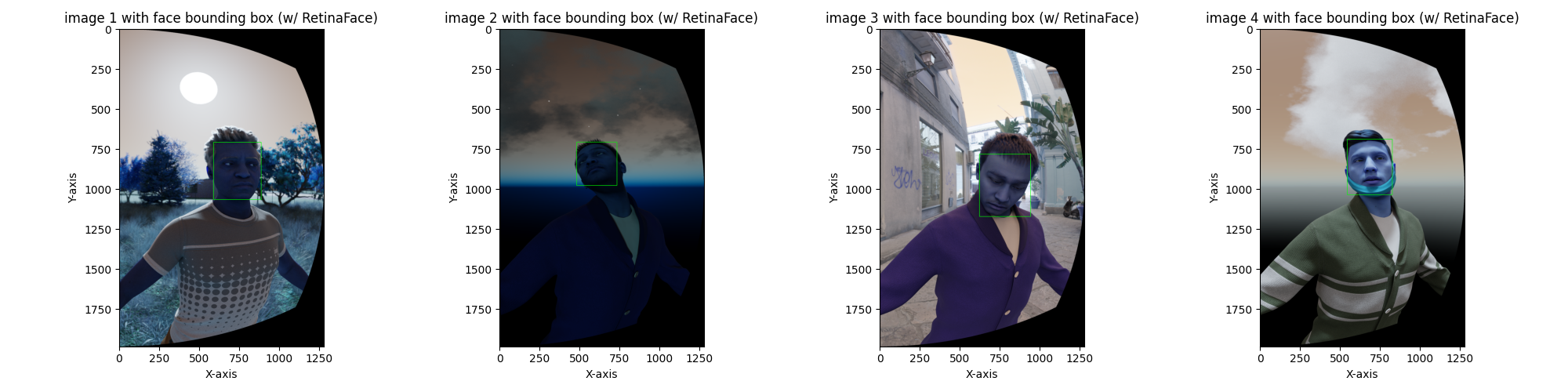}
    \caption{Image samples from the SynthC dataset. The RetinaFace pre-trained model~\cite{serengil2020lightface, serengil2021lightface} has been applied to extract the face bounding box (i.e., green box) of the driver.}
    \label{fig:Datagen_sample_faces_retina}
\end{figure}

\paragraph{SynthC:}
The SynthC dataset is comprised of 1,920 images featuring humans in diverse settings, including both indoor and outdoor environments, during different times of the day. Unlike the other datasets, the subjects in these images are not situated in a vehicle. While the diversity is higher than that of SynthA, it is not as extensive as SynthB. Samples of the SynthC dataset are demonstrated in Figure~\ref{fig:Datagen_sample_faces_retina}.

\paragraph{Label Annotations}
Each dataset comes with a rich set of labels and annotations. However, for the scope of this study, we are particularly interested in the labels related to facial attributes. These labels are categorized as follows:

\textbf{Gaze Labels:}
\begin{itemize}
    \item \textbf{SynthA}: 5 planes = \texttt{\{`HUD', `instrument\_cluster', `center\_console', `driver\_front\_windshield', \\ `middle\_front\_windshield'\}}
    
    \item \textbf{SynthB}: 5 planes = \texttt{\{`infotainment', `ext\_mirror', `int\_mirror', `rear', `road'\}}
    
    \item \textbf{SynthC}: 17 planes, re-annotated to 7 desired planes = \texttt{\{`center\_stack\_area', `road\_area', `up', `down', `passenger\_side', `driver\_side', `rearview\_mirror'\}}
\end{itemize}

\textbf{Age Labels:}
\begin{itemize}
    \item \textbf{SynthA}: Numerical values, re-annotated to 4 categories = \texttt{\{`13-18', `19-30', `31-50', `50+'\}}
    
    \item \textbf{SynthB}: 6 categories = \texttt{\{`0-3', `4-12', `13-18', `19-30', `31-50', `50+'\}}
    
    \item \textbf{SynthC}: 3 categories = \texttt{\{`young', `adult', `old'\}}
\end{itemize}

\textbf{Facial Expression Labels:}
\begin{itemize}
    \item \textbf{SynthA}: 3 categories = \texttt{\{`eyes\_closed', `compressed', `none'\}}
    
    \item \textbf{SynthB}: 6 categories = \texttt{\{`happy', `surprised', `angry', `random', `neutral', `sad'\}}
    
    \item \textbf{SynthC}: 9 categories = \texttt{\{`happiness', `disgust', `contempt', `fear', `surprise', `none', `anger', `mouth\_open', `sadness'\}}
\end{itemize}

These labels serve as the ground truth for training and evaluating our multi-task learning model. The diversity and richness of these labels across different datasets provide a robust platform for assessing the model's performance in various real-world-like scenarios.

\begin{figure}[t!]
    \centering
    \includegraphics[width=1\linewidth]{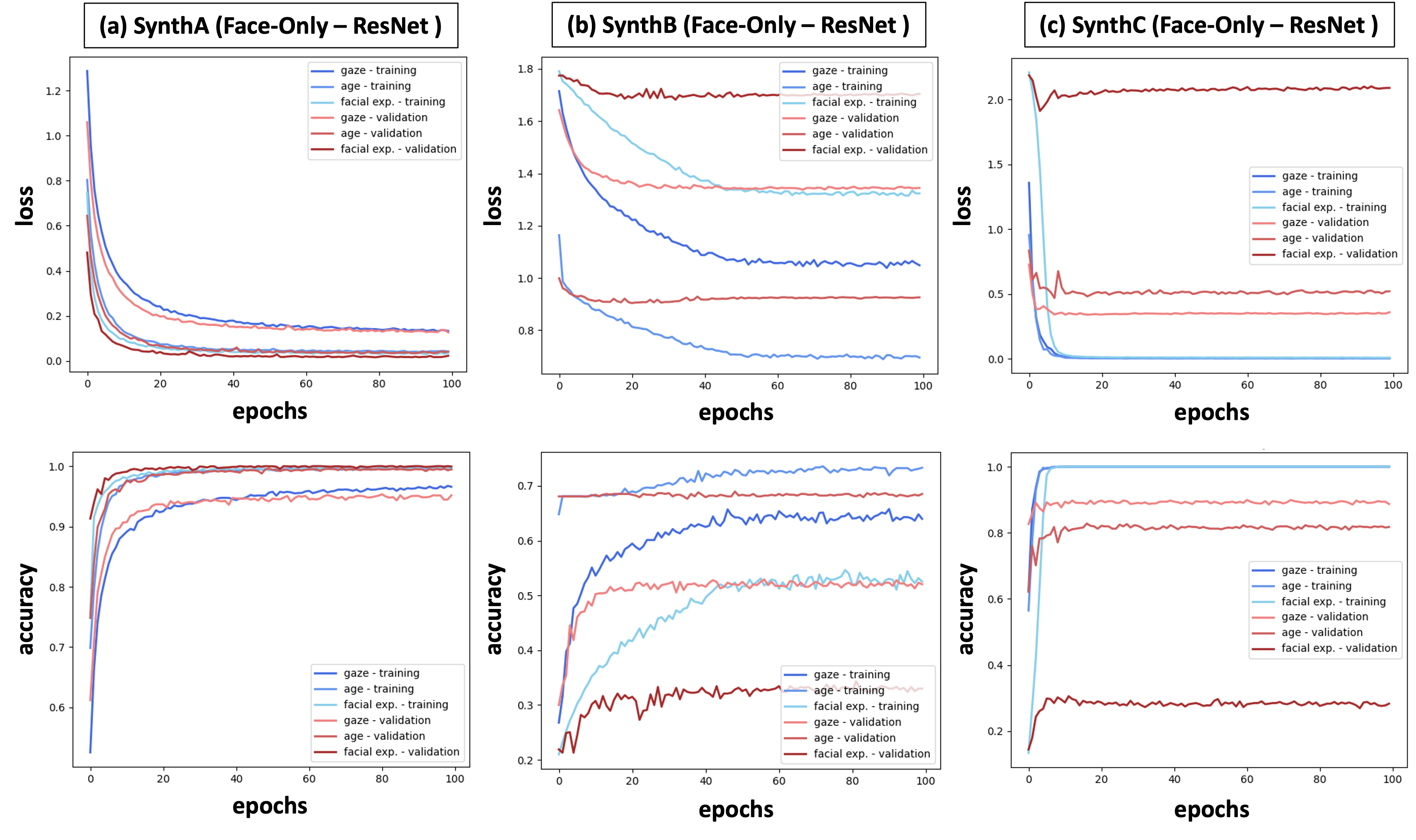}
    \caption{Performance of the pre-trained ResNet vision foundation model on the employed synthetic face-only datasets (i.e., (a) SynthA, (b) SynthB, and (c) SynthC) for in-vehicle multi-task facial attribute recognition. Top and bottom row demonstrate the task-specific losses and accuracies, respectively, for both training (shades of red) and validation (shades of blue) over epochs. All results use the FFT adaptation method.}
    \label{fig:faceonly_resnet_all}
\end{figure}

\subsection{Model Performances}
\label{subsec:ModelPerformances}
In this subsection, we delve into the performance metrics of the multi-task learning models built on pre-trained Vision Transformer (ViT) and ResNet architectures. We present comprehensive results on the face-only (Section~\ref{subsec:Preprocessing}) synthetic data, including learning curves for both training and evaluation phases. These curves encapsulate total loss, accuracy, and task-specific metrics for each synthetic dataset and for both vision foundation model architectures for in-vehicle multi-task facial attribute recognition.
\begin{figure}[t!]
    \centering
    \includegraphics[width=1\linewidth]{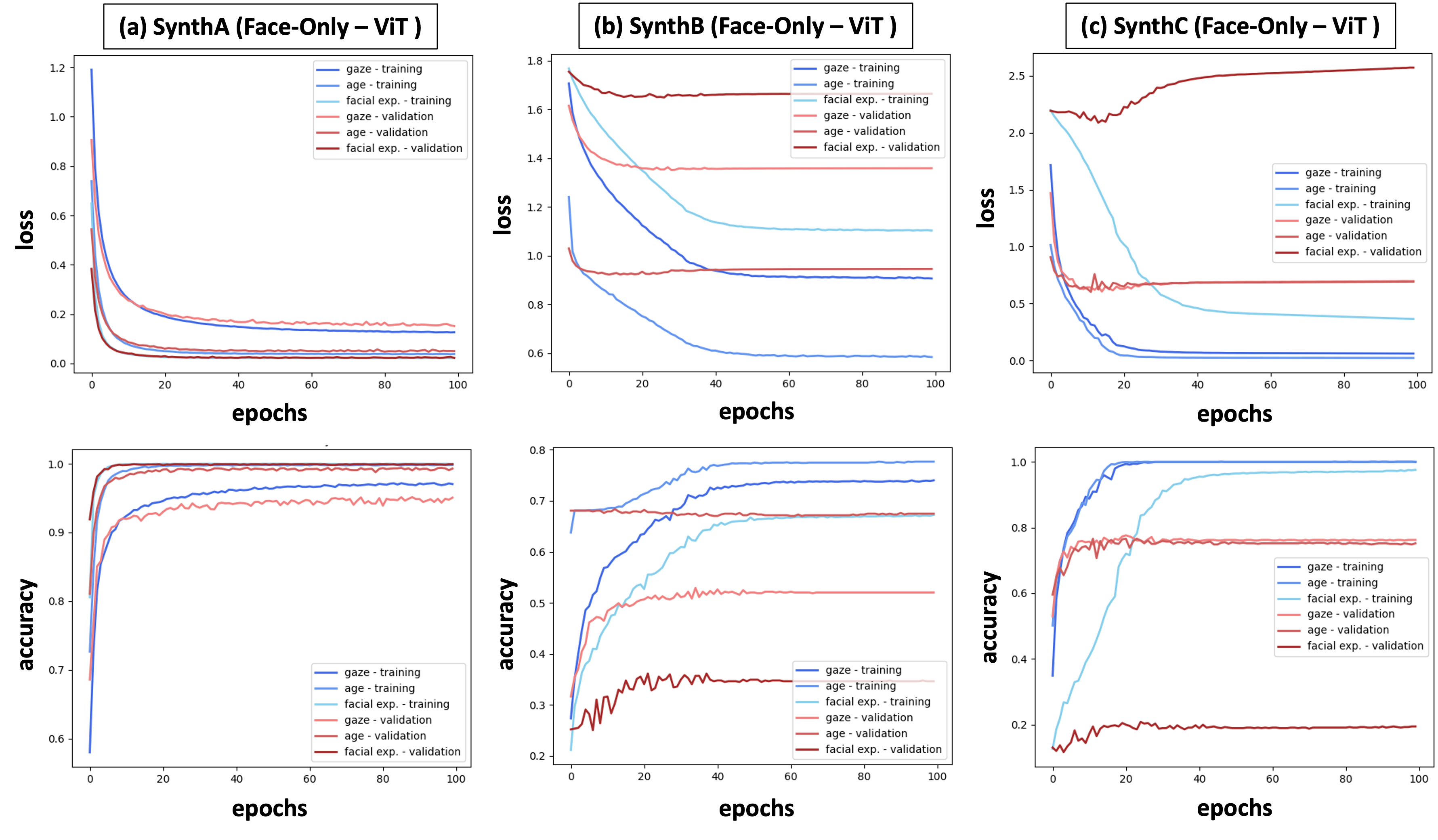}
    \caption{Performance of the pre-trained Vision Transformer (ViT) vision foundation model on the employed synthetic face-only datasets (i.e., (a) SynthA, (b) SynthB, and (c) SynthC) for in-vehicle multi-task facial attribute recognition. Top and bottom row demonstrate the task-specific losses and accuracies, respectively, for both training (shades of red) and validation (shades of blue) over epochs. All results use the prefix tuning adaptation method.}
    \label{fig:faceonly_vit_all}
\end{figure}

\subsubsection{Performance on Individual Datasets}
\label{subsubsec:ModelPerformances_Individual}
Figure~\ref{fig:faceonly_resnet_all} shows the performance of the pre-trained ResNet vision foundation model on the employed synthetic face-only datasets (i.e., (a) SynthA, (b) SynthB, and (c) SynthC) for in-vehicle multi-task facial attribute recognition. Top and bottom row demonstrate the task-specific losses and accuracies, respectively, for both training (shades of red) and validation (shades of blue) over epochs. All results use the FFT adaptation method. As our results indicate, the pre-trained ResNet model excels on the SynthA dataset, achieving near-optimal in-distribution performance. We attribute this high level of performance to the dataset's large volume and relatively low diversity compared to the other datasets used. In contrast, the SynthC dataset, which contains only 1,920 samples—a number generally considered insufficient for computer vision tasks—still yields impressive results when using the pre-trained ResNet model, outperforming the SynthB dataset. It's worth noting that the facial expression recognition task on the SynthC dataset involves classifying among nine different categories, adding complexity to an already challenging task due to the limited dataset size. Despite these constraints and the data-intensive nature of multi-task learning, our findings underscore the efficacy of transfer learning and the robustness of pre-trained foundation models in such scenarios.

Figure~\ref{fig:faceonly_vit_all} shows the performance of the pre-trained Vision Transformer (ViT) vision foundation model on the employed synthetic face-only datasets (i.e., (a) SynthA, (b) SynthB, and (c) SynthC) for in-vehicle multi-task facial attribute recognition. Top and bottom row demonstrate the task-specific losses and accuracies, respectively, for both training (shades of red) and validation (shades of blue) over epochs. All results use the prefix tuning adaptation method. As evidenced by our results, the pre-trained ViT model also attains near-optimal performance on the SynthA dataset (in-distribution), mirroring the performance trend observed with the ResNet model. However, a comparative analysis of Figures~\ref{fig:faceonly_resnet_all} and \ref{fig:faceonly_vit_all} reveals that the ResNet model outperforms the ViT model in several cases. Specifically, ResNet demonstrates superior task accuracies, enhanced data efficiency, and faster convergence rates, particularly when applied to the SynthC dataset. This result is particularly surprising given the recent successes of ViT architectures in various vision tasks. We hypothesize that the feature space learned by ViT may be overly complex for our specific multi-task problem, whereas the feature space of ResNet appears to be more amenable to the tasks at hand.

\subsubsection{Performance on Combined Dataset}
\label{subsubsec:ModelPerformances_Combined}
For the combined dataset, we utilized the face-only data extracted from the original datasets. This resulted in a dataset comprising 14,919 total samples and a more diverse data distribution. Note that, given the disparate label annotations across the original datasets, we re-annotated the labels to create a unified set. The re-annotated labels are as follows:
\begin{itemize}
    \item \textbf{Gaze Labels}: 6 planes \texttt{\{`infotainment', `ext\_mirror', `int\_mirror', `rear', `road', `passenger'\}}
    \item \textbf{Age Labels}: 3 categories \texttt{\{`teen', `adult', `elderly'\}}
    \item \textbf{Facial Expression Labels}: 5 categories \texttt{\{`happy', `surprised', `frown', `neutral', `sad'\}}
\end{itemize}

\begin{figure}[t!]
    \centering
    \includegraphics[width=1\linewidth]{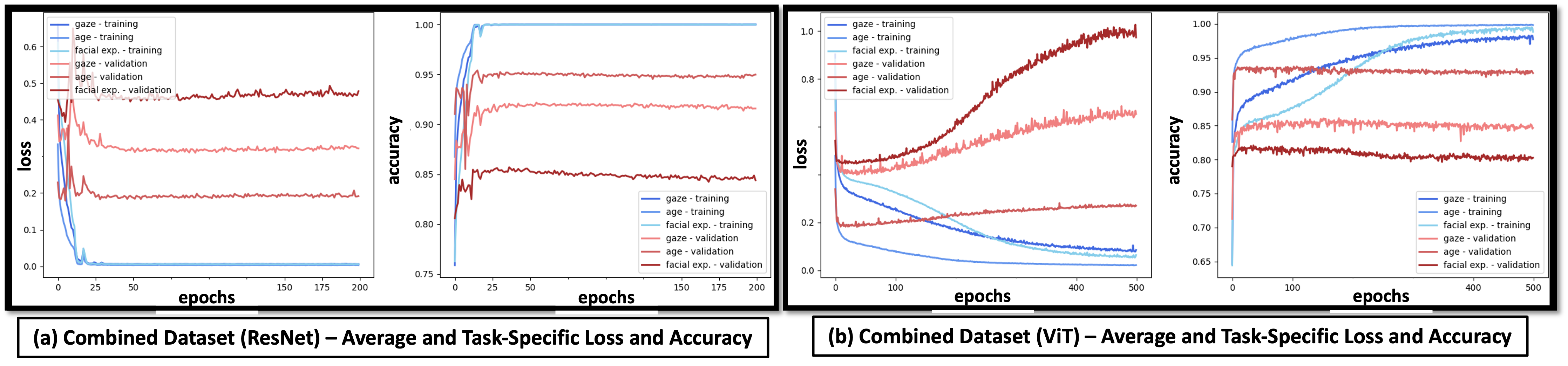}
    \caption{Figures (a) and (b) show the performance of the pre-trained Vision Transformer (ViT) and ResNet vision foundation models on the combined face-only datasets, respectively, for in-vehicle multi-task facial attribute recognition. In each figure, the left and right plot demonstrate the task-specific losses and accuracies, respectively, for both training (shades of red) and validation (shades of blue) over epochs.}
    \label{fig:combined_results_all}
\end{figure}

Figures (a) and (b) in Fig.~\ref{fig:combined_results_all} show the performance of the pre-trained ViT and ResNet foundation models on the combined face-only datasets, respectively, for in-vehicle multi-task facial attribute recognition. In each figure, the left and right plot demonstrate the task-specific losses and accuracies, respectively, for both training (shades of red) and validation (shades of blue) over epochs. As illustrated in our findings, both the pre-trained ResNet and ViT models exhibit significant performance improvements when trained on the aggregated synthetic dataset, relative to their performance on individual datasets as discussed in Section~\ref{subsubsec:ModelPerformances_Individual}. This uptick in performance can be attributed to the augmented data volume and a more diverse data distribution, which collectively contribute to a more robust and generalizable model compared to when trained on individual datasets. Additionally, once again, it is noteworthy that the pre-trained ResNet model consistently outperforms the ViT model, achieving not only higher accuracy rates but also more rapid convergence. Moreover, the ViT model exhibits a tendency to overfit (increasing validation loss), despite the implementation of several regularization techniques (e.g., loss regularization and dropout layers).

\subsection{Ablation Studies}
\label{subsec:Ablation_Studies}
In this section, we present a comprehensive ablation study to dissect the performance of our multi-task models. We focus on two primary aspects: the comparison between full-image and face-only data combinations, and the impact of curriculum learning on improving model performance.

\subsubsection{Study 1: Comparing Foundation Models and Preprocessing Steps}
\label{subsubsec:FullImageVSFaceOnly}
In this ablation study, we aim to rigorously evaluate the performance of the two foundation models used in our study across different preprocessing conditions. Specifically, we compare the efficacy of these models on the full-image individual datasets against the face-only individual datasets (presented in Figures~\ref{fig:faceonly_resnet_all} and \ref{fig:faceonly_vit_all}). The latter involves an additional face-extraction preprocessing step, as described in Section~\ref{subsec:Preprocessing}. This comparison serves to elucidate the impact of preprocessing techniques alongside the foundation model choice on the multi-task facial attribute recognition tasks, thereby providing insights into the optimal configurations for in-vehicle perception systems.
\begin{figure}[t!]
    \centering
    \includegraphics[width=1\linewidth]{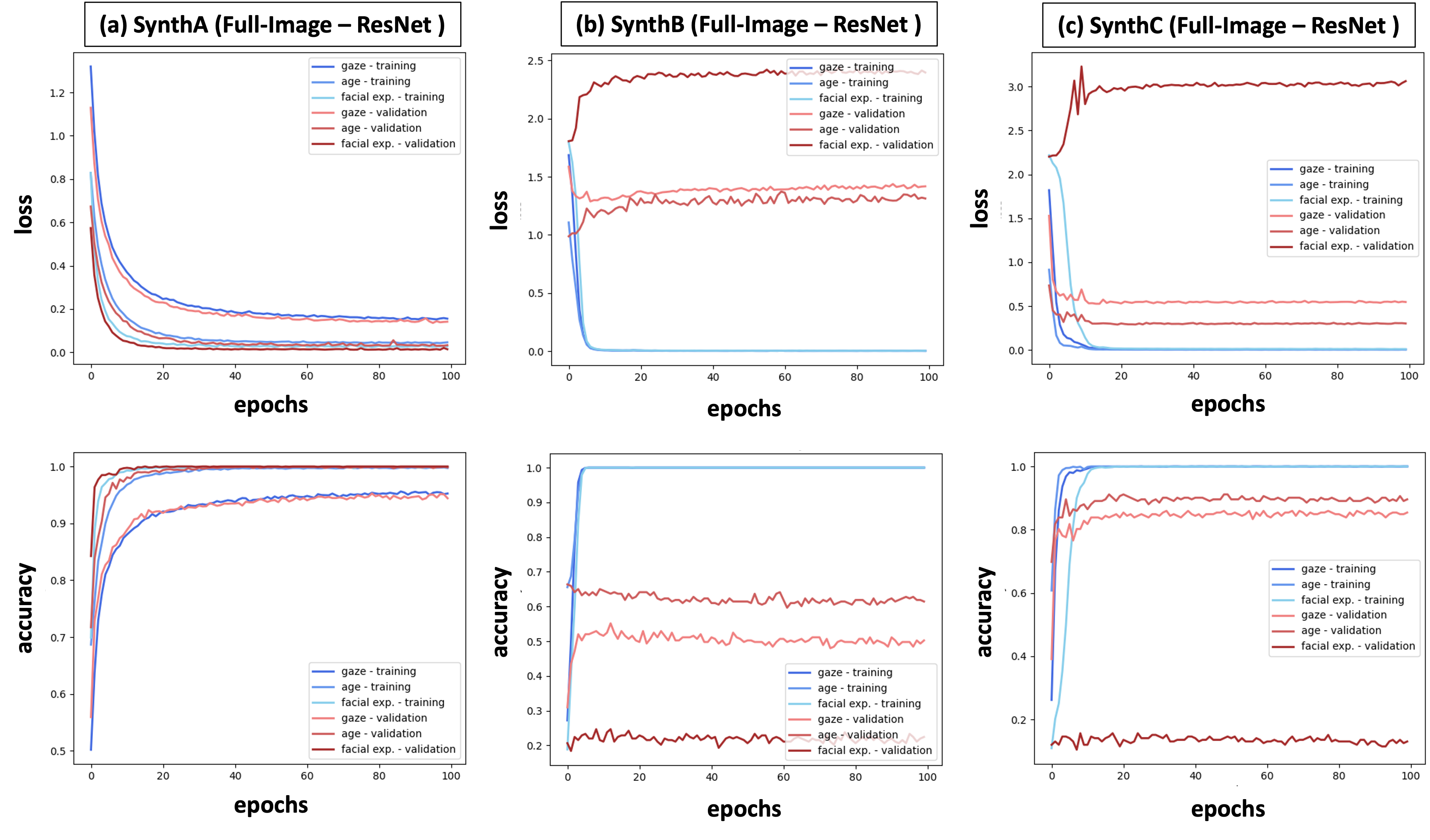}
    \caption{Performance of the pre-trained ResNet vision foundation model on the employed full-image (i.e., no face extraction in preprocessing) synthetic datasets (i.e., (a) SynthA, (b) SynthB, and (c) SynthC) for in-vehicle multi-task facial attribute recognition. Top and bottom row demonstrate the task-specific losses and accuracies, respectively, for both training (shades of red) and validation (shades of blue) over epochs. All results use the FFT adaptation method.}
    \label{fig:fullimage_resnet_all}
\end{figure}

\begin{figure}[t!]
    \centering
    \includegraphics[width=1\linewidth]{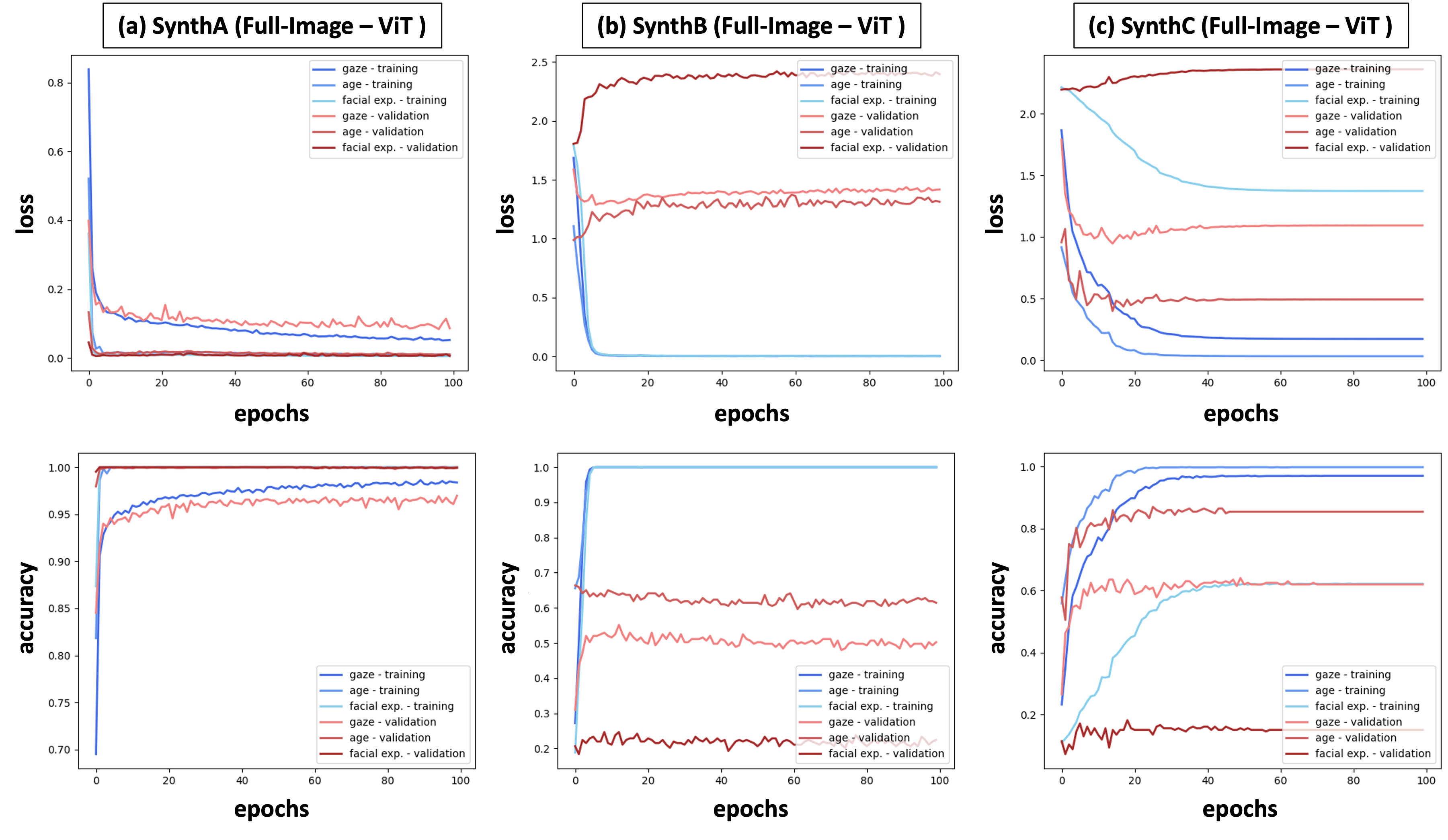}
    \caption{Performance of the pre-trained sViT vision foundation model on the employed full-image (i.e., no face extraction in preprocessing) synthetic datasets (i.e., (a) SynthA, (b) SynthB, and (c) SynthC) for in-vehicle multi-task facial attribute recognition. Top and bottom row demonstrate the task-specific losses and accuracies, respectively, for both training (shades of red) and validation (shades of blue) over epochs. All results use the PT adaptation method.}
    \label{fig:fullimage_vit_all}
    \vspace*{-0.5cm}
\end{figure}

Figures~\ref{fig:fullimage_resnet_all} and \ref{fig:fullimage_vit_all} demonstrate the performances of the pre-trained ResNet and ViT foundation models on the employed full-image synthetic datasets (i.e., (a) SynthA, (b) SynthB, and (c) SynthC), respectively. Top and bottom row demonstrate the task-specific losses and accuracies, respectively, for both training (shades of red) and validation (shades of blue) over epochs. All results use the FFT adaptation method. In summary, our results indicate that models trained on face-only data generally outperform those trained on full-image data. This suggests that focusing on the region of interest (i.e., the face) can lead to more accurate and reliable predictions. Additionally, we found that the ResNet foundation model, when fully fine-tuned, generally outperforms the ViT model with prefix tuning. This is noteworthy as full fine-tuning of ViT models is computationally expensive and often impractical.

\subsubsection{Study 2: Comparing Adaptation Methods}
\label{subsubsec:ComparingAdaptationMethods}
In this part of the ablation study, we focus on investigating the performance implications of various adaptation methods, i.e., Linear Probing (LP), Prefix Tuning (PT), and Full Fine Tuning (FFT), on each of the three individual synthetic datasets. The objective is to understand how different adaptation strategies affect the model's ability to generalize and perform well on multi-task facial attribute recognition, particularly when trained on limited synthetic data. This comparative analysis aims to identify the most effective adaptation method for each synthetic dataset, thereby offering valuable insights for model deployment in real-world applications. For this experiment we only use the pre-trained ResNet model as it achieved the best individual and aggregated dataset performances.

\begin{figure}[t!]
    \centering
    \includegraphics[width=1\linewidth]{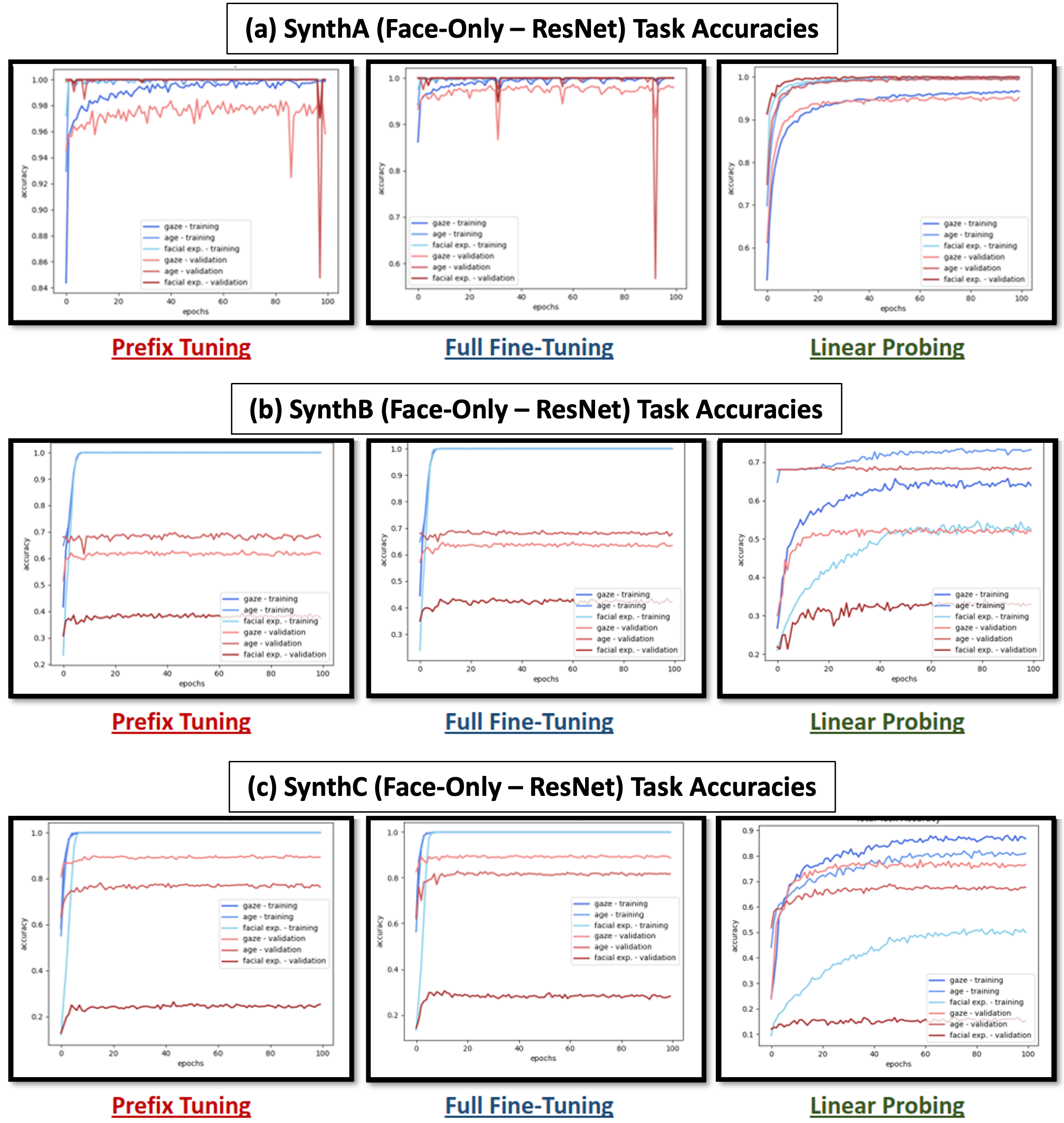}
    \caption{Performance (task-specific accuracies) of various adaptation methods, i.e., Linear Probing (LP), Prefix Tuning (PT), and Full Fine Tuning (FFT), on each of the three individual synthetic datasets.}
    \label{fig:adaptation_ablations_all}
\end{figure}

Figure~\ref{fig:adaptation_ablations_all} presents the task-specific accuracies attained by the pre-trained ResNet model using different adaptation techniques—Linear Probing (LP), Prefix Tuning (PT), and Full Fine Tuning (FFT)—across each of the three individual synthetic datasets. Remarkably, FFT consistently outperforms both LP and PT, a result that is particularly impressive given the limited dataset sizes. This underscores the adaptability and efficacy of pre-trained models in tackling new tasks. As anticipated, LP exhibits the slowest adaptation rate, primarily because only the task-specific heads are fine-tuned, while the foundational layers responsible for feature extraction remain static.

\subsubsection{Ablation Results Summary}
\label{subsec:Ablation_Results_Summary}
In this section we present two summary tables (Tables~\ref{tab:ablation_study_full_image} and \ref{tab:ablation_study_face_only}) that encapsulate our overall results and ablation study findings. Tables~\ref{tab:ablation_study_full_image} and \ref{tab:ablation_study_face_only} below show the training (T) and validation (V) accuracies for each task (gaze, age, facial expression) and for both face-only and full-image data, respectively.
\begin{table}[h!]
\centering
\caption{Ablation Study Results for Face-Only Dataset. The table presents the best training (T) and validation (V) results achieved via the best model.}
\label{tab:ablation_study_face_only}
\resizebox{\textwidth}{!}{%
\begin{tabular}{|c||c|c|c|c||p{3cm}|}
\hline
\multicolumn{1}{|c||}{\centering Foundation Model} & Dataset & Gaze (T -- V) & Age (T -- V) & Facial Exp. (T -- V) & \multicolumn{1}{c|}{\centering Adaptation Method} \\
\hline
\hline
\multirow{4}{*}{ViT} & SynthA & T: 100\% -- V: 95.0\% & T: 100\% -- V: 98.5\% & T: 100\% -- V: 100\% & Linear Probing \\
\cline{2-6}
& SynthB & T: 100\% -- V: 58.5\% & T: 100\% -- V: 69.0\% & T: 100\% -- V: 39.8\% & Prefix Tuning \\
\cline{2-6}
& SynthC & T: 100\% -- V: 86.5\% & T: 100\% -- V: 76.2\% & T: 100\% -- V: 32.7\% & Prefix Tuning \\
\cline{2-6}
& \textbf{Combined} & \textbf{T: 100\% -- V: 85.1\%} & \textbf{T: 97.5\% -- V: 94.2\%} & \textbf{T: 99.5\% -- V: 82.0\%} & \textbf{Prefix Tuning} \\
\hline
\hline
\multirow{4}{*}{ResNet} & SynthA & T: 100\% -- V: 98\% & T: 100\% -- V: 100\% & T: 100\% -- V: 100\% & Full Fine Tuning \\
\cline{2-6}
& SynthB & T: 100\% -- V: 64.5\% & T: 100\% -- V: 70.0\% & T: 100\% -- V: 43.7\% & Full Fine Tuning \\
\cline{2-6}
& SynthC & T: 100\% -- V: 90.0\% & T: 100\% -- V: 85.0\% & T: 100\% -- V: 34.2\% & Full Fine Tuning \\
\cline{2-6}
& \textbf{Combined} & \textbf{T: 100\% -- V: 93.1\%} & \textbf{T: 100\% -- V: 95.5\%} & \textbf{T: 100\% -- V: 86.0\%} & \textbf{Full Fine Tuning} \\
\hline
\end{tabular}%
}
\end{table}

\begin{table}[h!]
\centering
\caption{Ablation Study Results for Full-Image Dataset. The table presents the best training (T) and validation (V) results achieved via the best model.}
\label{tab:ablation_study_full_image}
\resizebox{\textwidth}{!}{%
\begin{tabular}{|c||c|c|c|c||p{3cm}|}
\hline
\multicolumn{1}{|c||}{\centering Foundation Model} & Dataset & Gaze (T -- V) & Age (T -- V) & Facial Exp. (T -- V) & \multicolumn{1}{c|}{\centering Adaptation Method} \\
\hline
\hline
\multirow{3}{*}{ViT} & SynthA & T: 100\% -- V: 97.8\% & T: 100\% -- V: 100\% & T: 100\% -- V: 100\% & Linear Probing \\
\cline{2-6}
& SynthB & T: 100\% -- V: 43.5\% & T: 100\% -- V: 64.2\% & T: 100\% -- V: 25.0\% & Prefix Tuning \\
\cline{2-6}
& SynthC & T: 98.1\% -- V: 61.1\% & T: 100\% -- V: 85.0\% & T: 61.0\% -- V: 19.9\% & Prefix Tuning \\
\hline
\hline
\multirow{3}{*}{ResNet} & SynthA & T: 100\% -- V: 99\% & T: 100\% -- V: 100\% & T: 100\% -- V: 100\% & Full Fine Tuning \\
\cline{2-6}
& SynthB & T: 100\% -- V: 55.5\% & T: 100\% -- V: 67.2\% & T: 100\% -- V: 26.5\% & Full Fine Tuning \\
\cline{2-6}
& SynthC & T: 100\% -- V: 87.1\% & T: 100\% -- V: 91.1\% & T: 100\% -- V: 18.5\% & Full Fine Tuning \\
\hline
\end{tabular}%
}
\end{table}

\subsubsection{Study 3: Training Techniques (Curriculum Learning)}
\label{subsubsec:CurriculumLearning}
In addition to the above comparisons, we also evaluated the impact of curriculum learning on model performance. In the context of MTL, curriculum learning serves as a strategic framework for model training that aims to improve the efficiency of the learning process. Traditional MTL often involves training a model on multiple tasks simultaneously, treating all tasks as equally important. However, curriculum learning introduces a pedagogical approach, akin to human learning, where tasks are organized in a sequence from simpler to more complex. Our findings suggest that while curriculum learning does not significantly improve the overall performance, it does lead to minor faster convergence.


\subsection{Out-of-Distribution Inference Results}
\label{subsec:OOD_Results}
Out-of-Distribution (OOD) inference serves as a pivotal evaluation metric for assessing the generalization capability of machine learning models, particularly in real-world scenarios where the model is likely to encounter data that diverges from the training set. This form of evaluation is especially crucial for synthetic datasets, which often excel in in-distribution performance but may falter when exposed to real-world, diverse data. In this subsection, we present an exhaustive analysis of the OOD performance of our best-performing models, i.e., multi-task facial attribute recognition model built on pre-trained ResNet foundation model, trained on aggregated face-only synthetic data and adapted via the FFT adaptation technique. specifically focusing on their ability to generalize to the Kaggle UTKFace dataset~\cite{zhang2017age}, a dataset that is significantly different in nature from the synthetic datasets used for training.
\begin{figure}[h!]
    \centering
    \includegraphics[width=1\linewidth]{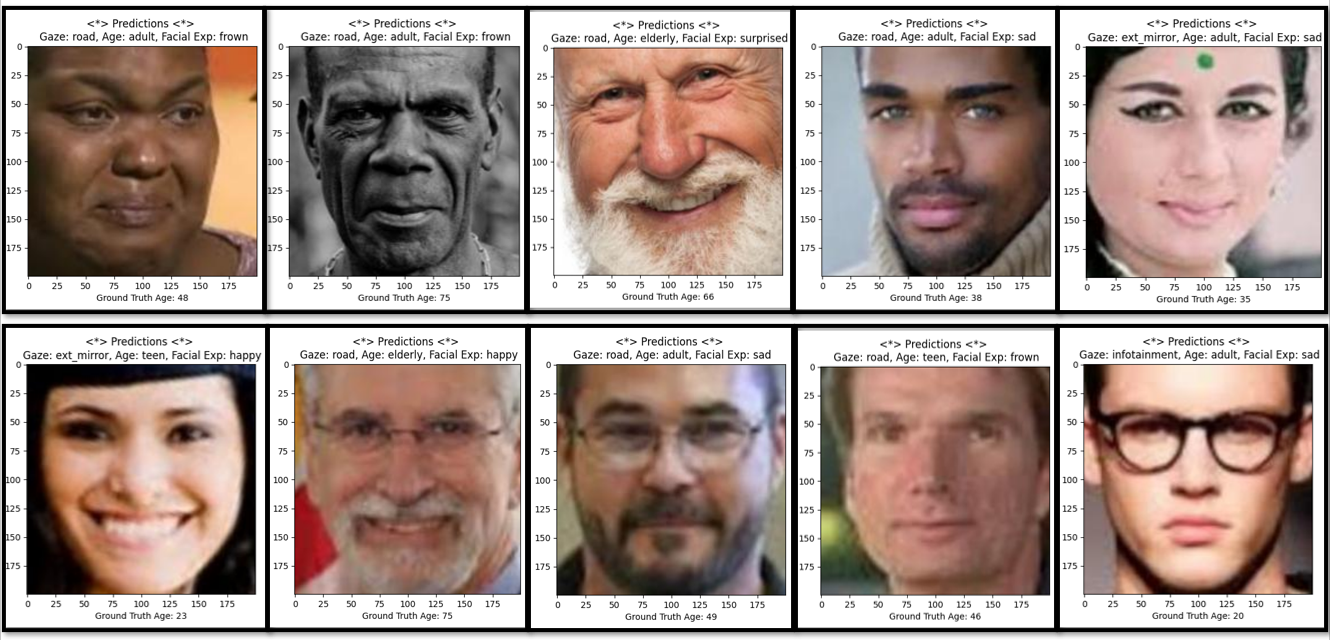}
    \caption{Out-of-Distribution (OOD) inference performance of our best model (i.e., multi-task facial attribute recognition model built on pre-trained ResNet foundation model, trained on aggregated face-only synthetic data and adapted via the FFT adaptation technique) obtained from the Kaggle UTKFace dataset (only ground truth age labels are provided by the dataset). Empirically, this model achieves an average accuracy of approximately 50\% across tasks on the assessed OOD data. These results underscore the remarkable potential of leveraging existing foundation models and synthetic data to construct robust and capable models.}
    \label{fig:OOD_Test}
\end{figure}

OOD inference was rigorously performed on the Kaggle UTKFace dataset, a dataset that was intentionally kept separate and was not seen by the model during any phase of the training process. This approach allows us to critically evaluate how well the model can generalize to new, unseen data, thereby providing a robust measure of its real-world applicability. We performed several OOD experiments specific to each of the datasets separately in order to characterize model performance with respect to data distribution and volume. In summary:
\begin{enumerate}
    \item \textbf{SynthA}: This dataset proved to be ``too synthetic" for real-world applications. The models trained on this dataset, while achieving remarkable near-optimal performance on in-distribution data, performed very poorly on OOD tasks, averaging less than 10\% accuracy (empirically) across all tasks. This suggests that while the dataset may be suitable for in-distribution tasks, it is not robust enough for real-world applications.
    
    \item \textbf{SynthB}: Models trained on this dataset performed better in OOD tasks compared to those trained on SynthA. However, the overall performance was still noticeably suboptimal, likely due to the simultaneous low sample size and wide data distribution.
    
    \item \textbf{SynthC}: Despite having the lowest sample size (1920 samples), models trained on this dataset outperformed the others in OOD tasks, averaging around 35\% accuracy (empirically). This suggests that SynthC provides a more balanced and robust training set for real-world applications.
\end{enumerate}

Finally, we also performed an OOD experiment on the models trained via the aggregated dataset. Combining the three datasets increases data diversity and data count which in theory should improve the performance in OOD inference. This is exactly what we observe in practice. Figure~\ref{fig:OOD_Test} shows the OOD inference performance of our best model (i.e., multi-task facial attribute recognition model built on pre-trained ResNet foundation model, trained on aggregated face-only synthetic data and adapted via the FFT adaptation technique) obtained from the Kaggle UTKFace dataset (only ground truth age labels are provided by the dataset). Empirically, this model achieves an average accuracy of approximately 50\% across tasks on the assessed OOD data. These results underscore the remarkable potential of leveraging existing foundation models and synthetic data to construct robust and capable models.

Nevertheless, to shed further light onto these OOD performances, a post-evaluation data distribution analysis is necessary (Section~\ref{subsec:Discussion}). For instance, we hypothesize that the aggregated dataset has a relatively bad label distribution given the re-annotation process that was required to merge the data samples. This bad distribution is almost unavoidable due to different nature of the three datasets.

\subsection{Discussion}
\label{subsec:small_discussion}
The synthetic nature of the data used for training has profound implications on the model's OOD performance, a critical factor for the deployment of machine learning models in real-world scenarios. Synthetic data, while highly controlled and easily customizable, often lacks the inherent variability and noise present in real-world data. This discrepancy can lead to models that perform exceptionally well on in-distribution data but falter when faced with real-world, diverse data (e.e., the SynthA dataset). Therefore, for synthetic data to be genuinely useful in real-world applications, it must be designed to be as diverse as possible, capturing a wide range of scenarios, conditions, and edge cases. Introducing controlled noise into the synthetic data can also be beneficial, as it can simulate the kind of data irregularities that a model is likely to encounter in a real-world setting.

In light of these challenges, the use of advanced techniques becomes indispensable for improving OOD performance. In this study, we employed a variety of such techniques, including data augmentation, fine-tuning and regularization methods, as well as transfer learning, adaptation, and robust evaluation metrics, to ensure that the models are not just memorizing the training data but are learning to generalize across different distributions. These advanced techniques serve as a compensatory mechanism for the limitations of synthetic data, enabling the model to bridge the gap between synthetic and real-world data. By doing so, we created multi-task models that are not only high-performing but also robust and reliable when deployed in real-world applications, thereby addressing one of the most significant challenges in the utilization of synthetic data for machine learning.

\subsection{Post-Evaluation Analysis: Data Distributions and Similarity}
\label{subsec:Discussion}
In this subsection, we delve into the data distributions of the synthetic datasets used for training the models. We investigate both image and label distributions to understand the characteristics of each dataset. This analysis aims to shed light on the observed model performances (in-distribution and out-of-distribution) and to provide insights into the limitations and potentials of each dataset.

\subsubsection{Visualizing Data Distributions and Similarity Metrics}
\label{subsubsec:Data_Label_Distributions}
We begin our analysis by presenting the data distributions in a structured manner. To achieve this, we first flatten the images, transforming them from multi-dimensional arrays into one-dimensional vectors. This step is crucial for simplifying the computational complexity involved in similarity calculations. We then compute two distinct metrics of similarity: Euclidean and Cosine similarities. These metrics serve different purposes; while Euclidean similarity measures the \textit{straight-line} distance between two points in the feature space, Cosine similarity assesses the angle between two vectors, providing insights into their directional relationship. To make these similarity measures comparable and to facilitate their interpretation, we then normalize these similarity matrices.

\begin{figure}[h!]
    \centering
    \begin{subfigure}{0.49\textwidth}
        \centering
        \includegraphics[width=\linewidth]{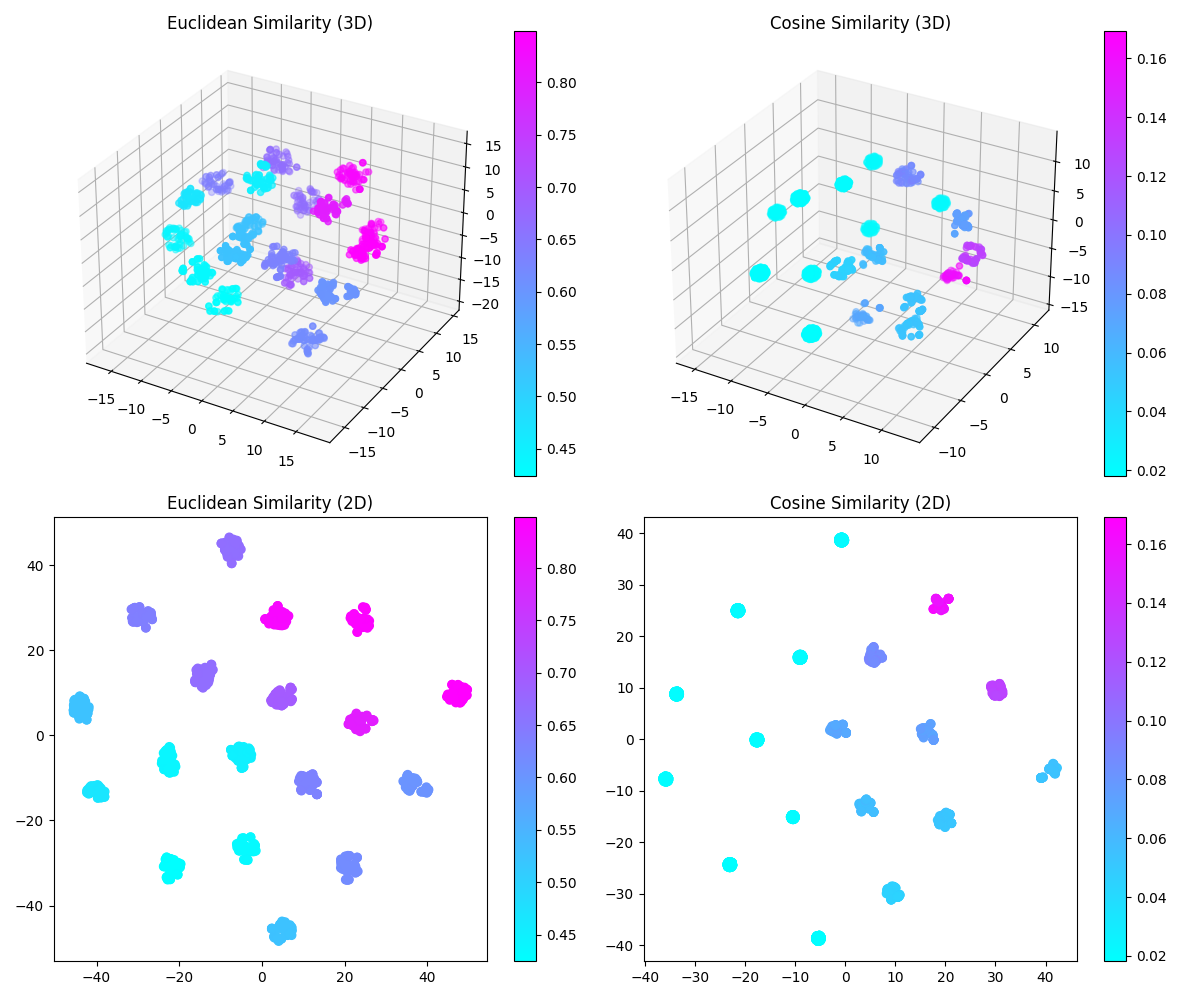}
        \caption{SynthA Data Distribution}
        \label{fig:Synthesis_data_dist_1000}
    \end{subfigure}%
    \begin{subfigure}{0.49\textwidth}
        \centering
        \includegraphics[width=\linewidth]{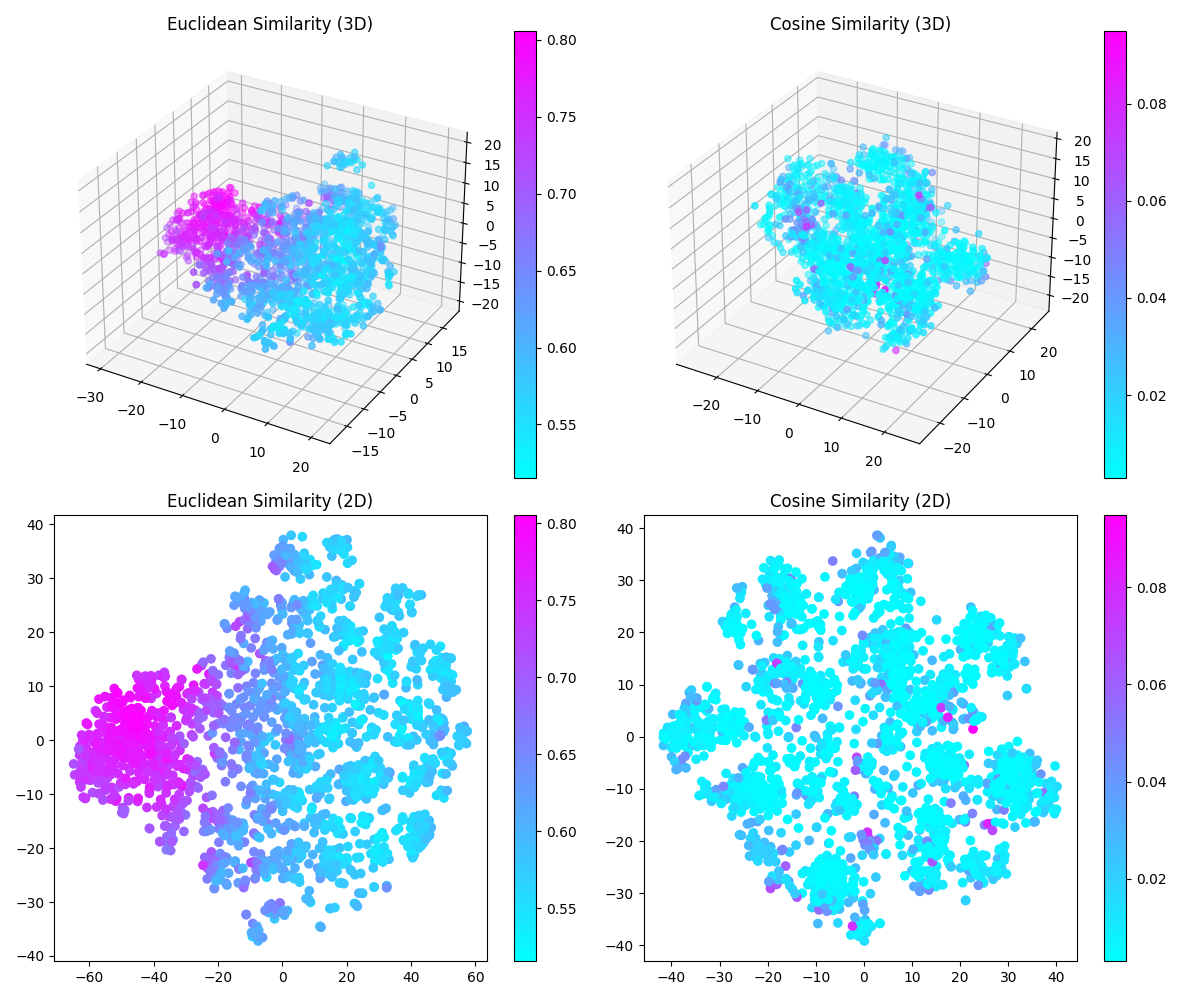}
        \caption{SynthB Data Distribution}
        \label{fig:Anyverse_data_distribution}
    \end{subfigure}
    \begin{subfigure}{0.49\textwidth}
        \centering
        \includegraphics[width=\linewidth]{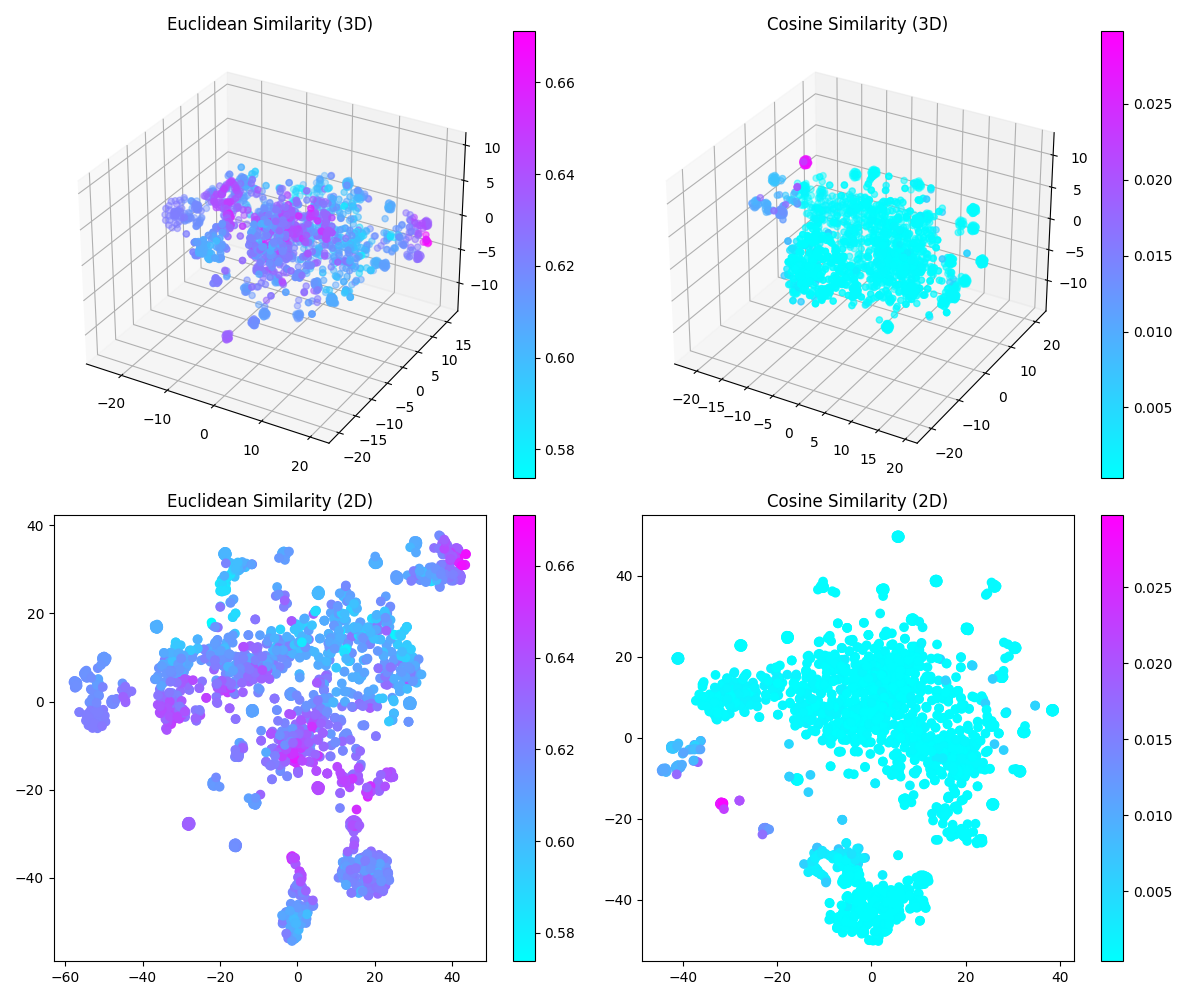}
        \caption{SynthC Data Distribution}
        \label{fig:Datagen_data_distribution}
    \end{subfigure}
    \caption{Data distribution (equal portions) of the three employed synthetic datasets. Each point in space represents an RGB image. Distances are computed via t-SNE algorithm and the color-maps represent two similarity metrics: (1) Euclidean similarity and (2) Cosine similarity. Both 2D and 3D representations are provided for better clarity. Refer to Section~\ref{subsubsec:UnderstandingAnalyzingScatterPlots} for details regarding understanding and interpreting the scatter plots.}
    \label{fig:data_distributions}
    \vspace*{-0.5cm}
\end{figure}

To further visualize the data distributions, we employ t-SNE (t-Distributed Stochastic Neighbor Embedding) for dimensionality reduction~\cite{cieslak2020t}. Each image, now represented as a flattened vector, is mapped to a point in a 2D or 3D space. These scatter plots are then color-coded based on the previously computed and normalized similarity matrices, providing a multi-faceted view of data relationships. It is worth noting that t-SNE is sensitive to the scale of the data. This sensitivity necessitates appropriate data preprocessing steps, such as normalization or standardization, to ensure that the t-SNE algorithm accurately captures the underlying data structure. Figures~\ref{fig:Synthesis_data_dist_1000}-\ref{fig:Datagen_data_distribution} present the data distribution (equal portions) of the three employed synthetic datasets. Each point in space represents an RGB image where the distances and color-maps are computed via t-SNE and the similarity metrics as described above. Refer to Section~\ref{subsubsec:UnderstandingAnalyzingScatterPlots} for a road-map regarding understanding and interpreting the scatter plots. The visualizations serve as a qualitative tool for exploring the relationships between images in the datasets. By examining clusters, outliers, and color mapping, and by comparing different similarity measures, we can gain valuable insights into the underlying structure of the data. These insights can guide further quantitative analysis, feature engineering, or model selection.

\subsubsection{Understanding and Analyzing the Data Distribution Scatter Plots}
\label{subsubsec:UnderstandingAnalyzingScatterPlots}
\begin{itemize}
    \item \textbf{Axes ($x, y, z$):} The axes represent the three main components obtained from the t-SNE algorithm. These components capture the most significant patterns in the data.
    \item \textbf{Color Mapping:} Colors represent either Euclidean or Cosine similarity. Lower Euclidean values and higher Cosine values indicate greater similarity.
    \item \textbf{Cluster Identification and Color Interpretation:} By looking for clusters of points in the 2D/3D spaces, we identify groups of similar images. We examine the color coding in conjunction with the spatial positioning and clustering behavior of the points.
    \item \textbf{Compare Euclidean vs. Cosine Similarity:} Differences between these plots could provide insights into the nature of the images.
    \item \textbf{Outlier Detection:} Points far away from others could represent unique or anomalous images.
\end{itemize}

\begin{figure}[h!]
    \centering
    \includegraphics[trim={0 0.75cm 0 0.5cm}, clip, width=1\linewidth]{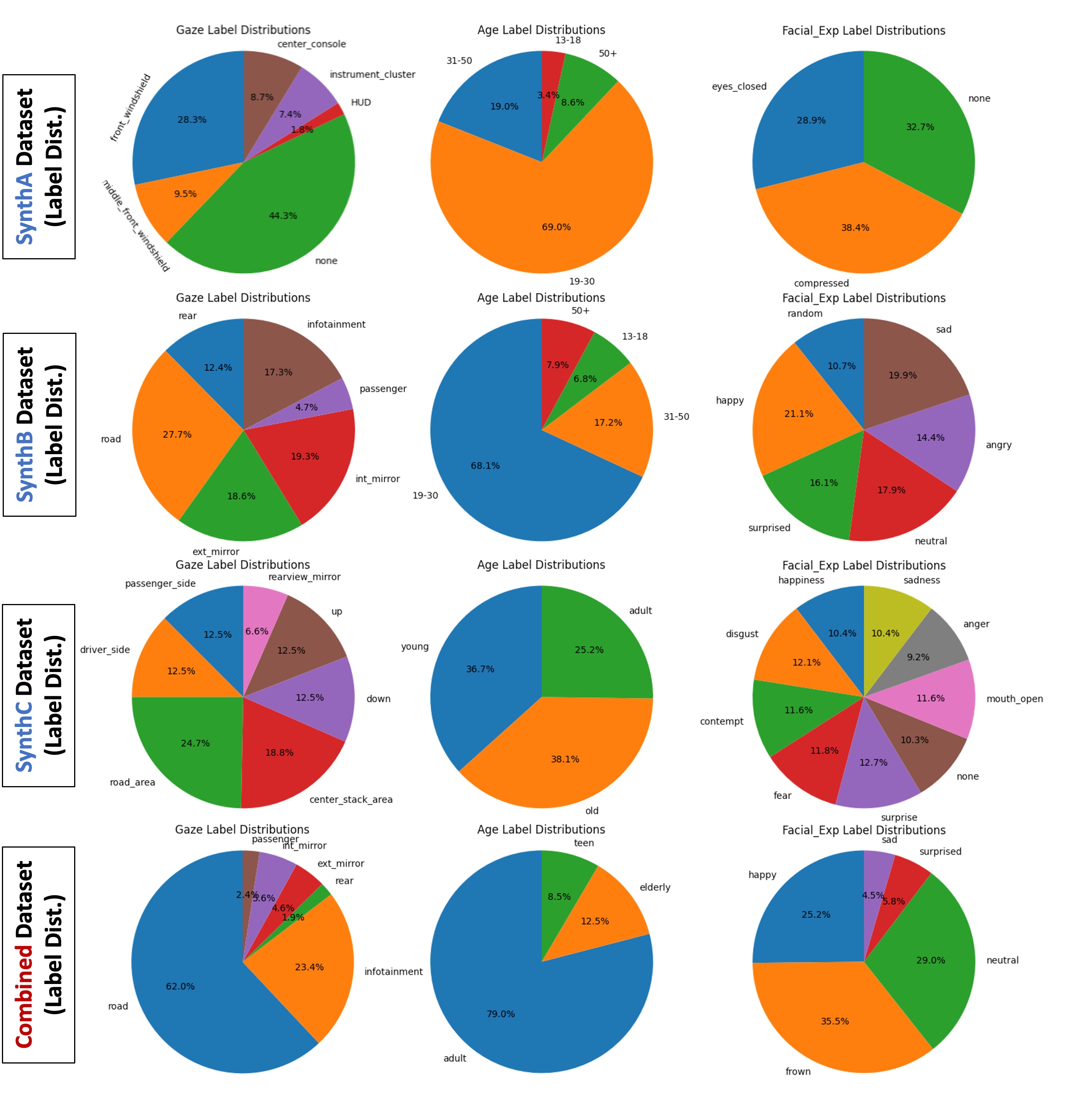}
    \caption{Label distributions across all three employed synthetic datastes, i.e., SynthA, SynthB, and SynthC, as well as the combined dataset.}
    \label{fig:label_dists_all}
\end{figure}

For label distributions, we count the number of labels for each task and represent this distribution as pie charts. This visualization helps in quickly grasping the balance or imbalance in the dataset for each task, which is essential for understanding model performance later on. Figure~\ref{fig:label_dists_all} shows the label distributions across all three employed synthetic datastes, i.e., SynthA, SynthB, and SynthC, as well as the combined dataset. See Section ~\ref{subsubsec:SummaryofDistributionandPerformanceObservations} for a summary of the data and label distribution observations.

\subsubsection{Summary of Data and Label Distribution Observations}
\label{subsubsec:SummaryofDistributionandPerformanceObservations}
\paragraph{Data Distribution Observations:}
\begin{enumerate}
    \item \textbf{SynthA}: Exhibits clear clustering behavior according to t-SNE, with each cluster having a distinct color (similarity metric).
    \item \textbf{SynthB}: Displays minor clustering and is highly dispersed in the feature space. The colors do not align with the clusters and vary significantly across the data distribution.
    \item \textbf{SynthC}: Similar to SynthB in terms of minor clustering but is less dispersed. The colors are generally similar for all data samples, unlike SynthB.
\end{enumerate}

\paragraph{Label Distribution Observations:}
\begin{enumerate}
    \item \textbf{SynthA}: The label distribution in this dataset is an example of a bad and unrealistic distribution for building a generalisable model. In two out of the three tasks (gaze and age), not only the label distribution is non-uniform, but also, more than half of the data have a similar label. This makes the high in-distribution performance of the model ``fake", as the model can only predict one label all the time and achieve 50\% or more accuracy.
    \item \textbf{SynthB}: Although the label distributions in this dataset is more uniform compared to SynthA, we still observe a large portion of the data labeled as \texttt{`19-30'} (i.e., 68\%) for the age classification task.
    \item \textbf{SynthC}: This dataset has the most uniform label distribution among the three synthetic datasets. Also, note that this dataset has most categories in two of the downstream classification tasks (gaze and facial expression).
    \item \textbf{Combined}: The aggregated dataset has a also a bad label distribution given the re-annotation process that was required to merge the data samples. This bad distribution is almost unavoidable due to different nature of the three datasets.
\end{enumerate}


\paragraph{Connections and Reasoning}
\begin{enumerate}
    \item \textbf{SynthA}: The clear clustering behavior in SynthA likely contributes to its excellent in-distribution performance. The model can easily learn the distinct clusters, leading to high accuracy. However, this also makes the model less generalizable, explaining its poor out-of-distribution performance.
    
    \item \textbf{SynthB}: The lack of clear clustering and the high dispersion in the feature space make it challenging for the model to learn meaningful patterns, leading to mediocre in-distribution performance. The varying similarity metrics across the data distribution further complicate the learning process, resulting in poor out-of-distribution performance.
    
    \item \textbf{SynthC}: The minor clustering and less dispersion in SynthC make it easier for the model to generalize, which is reflected in its decent in-distribution and out-of-distribution performance. The more uniform similarity metrics across the data samples likely aid in this generalization.
\end{enumerate}

In summary, the data distribution characteristics directly influence the model's ability to learn and generalize, which is evident from the performance observations on the three datasets.

\subsubsection{Summary of Post-Evaluation Analysis Findings}
\label{subsubsec:SummaryPost-EvaluationAnalysisFindings}
In this section, we distill the essential findings from our extensive experiments and post-evaluation analyses. These key findings serve as a concise summary that encapsulates the strengths and weaknesses of each dataset in terms of their utility for training robust machine learning models, particularly in the context of out-of-distribution performance.
\begin{enumerate}
    \item \textbf{SynthA:} This dataset is almost ``\textit{too synthetic}", which is evident from its near-perfect in-distribution performance. However, this perfection becomes its Achilles' heel when it comes to OOD performance. The models trained on this dataset struggle to generalize to real-world data, as evidenced by their poor OOD performance. The synthetic nature of the data seems to lack the diversity and noise inherent in real-world scenarios (Figure~\ref{fig:Synthesis_data_dist_1000}), making it less suitable for training models that need to operate effectively in less controlled environments.
    
    \item \textbf{SynthB:} This dataset stands out for its wide distribution, encompassing a broad range of features and characteristics. However, its utility is significantly hampered by its low sample size. Given that machine learning models, especially multi-task foundation models like the ones we are working with, are data-hungry, the low sample size becomes a limiting factor. The wide distribution coupled with a low sample size results in a dataset that is not representative enough for training robust models. This leads to poor generalization and suboptimal performance in both in-distribution and OOD scenarios.
    
    \item \textbf{SynthC:} Among the datasets examined, SynthC offers a more balanced distribution of features. It strikes a middle ground between the overly synthetic nature of SynthA and the wide but sparse distribution of SynthB (the more uniform similarity metrics across the data samples). This balance makes it a promising candidate for training more robust models that can generalize well to unseen data. While its sample size is still very low, the quality and distribution of the data make up for it to some extent, as evidenced by its relatively better in-distribution and OOD performance.

    \item \textbf{Combined:} Combining the three datasets increases data diversity and data count which in theory and practice improved the performance for both in-distribution data and OOD inference. This OOD results was observed despite the inevitable bad label distribution in aggregated data, which highlights the importance of data diversity and high data counts. To make the OOD performance even better using the combined dataset, we would need to increase counts for labels that have a low count (Figure~\ref{fig:label_dists_all}).
\end{enumerate}

\section{Conclusion}
\label{sec:DiscussionAndConclusion}
In this study, we have rigorously evaluated the performance of Vision Transformer (ViT) and ResNet foundation models on multi-task vision problems using synthetic datasets. We particularly focused on building robust and generalisable perception models in-vehicle multi-task facial attribute recognition. Our comprehensive experiments encompassed various aspects such as model architectures, data types, adaptation methods, and out-of-distribution performance analysis. Our best performing model was comprised of a multi-task facial attribute recognition model built on pre-trained ResNet foundation model, trained on aggregated face-only synthetic data and adapted via the FFT adaptation technique. Our results indicated the impressive capability of foundation models and transfer learning in learning multiple complex perception tasks, even when train on limited amount of synthetic data. The findings of this research demonstrate a great potential in benefiting from synthetic data where models were able to learn complex computer vision tasks within a few training epochs via employing transfer learning from existing vision foundation models. The results also revealed that while synthetic data can be highly effective for in-distribution tasks, they pose challenges for generalization to real-world and out-of-distribution scenarios. We also delved into the data distributions of the synthetic datasets, employing both image and label distributions to understand their characteristics. Through visualizations and quantitative analyses, we gained valuable insights into the limitations and potentials of each dataset.

\subsection{Key Research Takeaways:}
\begin{enumerate}
    \item \textbf{In-Vehicle Perception and Intelligence:} The significance of in-vehicle perception and intelligence cannot be overstated, especially when it comes to multi-task facial attribute recognition. This technology is pivotal for enhancing the safety and personalized experience of passengers, thereby making autonomous and semi-autonomous vehicles more reliable and user-friendly.
    
    \item \textbf{High-Quality Real-World Data:} To achieve the level of precision required for in-vehicle perception, it is crucial to have access to high-quality and well-distributed real-world data. The quality of the data directly influences the model's ability to generalize and perform reliably in diverse conditions, thereby reinforcing the importance of the first point.
    
    \item \textbf{Synthetic Data:} In the absence of sufficient real-world data, synthetic data serves as a viable alternative. However, the utility of synthetic data is not just a stopgap but a strategic asset that can simulate various edge cases, thereby aiding in the robustness of in-vehicle perception systems.
    
    \item \textbf{Vision Foundation Models:} Vision foundation models become particularly important when high-quality real-world data is scarce. These pre-trained models can be fine-tuned and adapted to perform specific tasks related to in-vehicle perception, thereby offering a quicker and more efficient route to achieving the goals set out in the first point.
    
    \item \textbf{Realistic Synthetic Data:} While synthetic data is beneficial, it needs to closely mimic real-world conditions in terms of distributions, noisiness, etc., to be truly effective. Data that is ``\textit{too synthetic}" can lead to poor out-of-distribution performance, undermining the real-world applicability of models designed for in-vehicle perception.
    
    \item \textbf{Future Research:} Conducting research like ours is vital for the future of the automotive industry and technology. It not only addresses immediate challenges in multi-task facial attribute recognition but also lays the groundwork for more advanced in-vehicle perception systems, thereby contributing to the broader goals of automotive safety and personalization. 
\end{enumerate}

\subsection{Limitations and Future Work}
\label{subsec:LimitationsAndFutureWork}
Despite the comprehensive nature of this study, there are limitations that warrant attention. First, the synthetic nature of the datasets used may not fully capture the complexities and variabilities of real-world data, thereby affecting the generalizability of the models. Second, the study was constrained by the sample sizes of the synthetic datasets, which could impact the robustness of the models and the reliability of the results. Future work should focus on incorporating more diverse and larger datasets to validate the findings of this study. Additionally, exploring other advanced techniques and architectures could provide further insights into improving both in-distribution and out-of-distribution performances. Finally, by employing model compression techniques~\cite{karimzadeh2022hardware, karimzadeh2022towards_2} such as network quantization~\cite{karimzadeh2022bits, karimzadeh2020hardware}, and model pruning~\cite{karimzadeh2020memory, karimzadeh2020hardware_2}, and hardware-aware design~\cite{karimzadeh2022towards, karimzadeh2022towards_2} we aim to further realize the potentials of such systems and models for in-vehicle perception and intelligence and real-world deployment.

\bibliographystyle{IEEEtran}
\bibliography{references}  






\end{document}